\documentclass[10pt,twocolumn]{fairmeta}

\newif\ifcvpr  
\cvprfalse

\usepackage{xspace}
\RequirePackage{amsmath}
\RequirePackage{amssymb}
\makeatletter
\DeclareRobustCommand\onedot{\futurelet\@let@token\@onedot}
\def\@onedot{\ifx\@let@token.\else.\null\fi\xspace}

\def\etc{\emph{etc}\onedot} 
\def\wrt{w.r.t\onedot}

\makeatother

\ifcvpr
\usepackage[table]{xcolor}
\fi 

\usepackage{multirow}
\usepackage{pifont}
\usepackage{sidecap}
\usepackage{colortbl}

\def\mypar#1{\vspace{1mm}{\noindent\bf #1.}\hspace{1mm}}

\newcommand{\ours}{DP-SIMS\xspace}

\newcommand{\cmark}{\ding{51}}%
\newcommand{\xmark}{\ding{55}}%

\newcommand{\ccc}[1]{{\bf\color{red}#1}}

\newcommand{\COCO}{COCO-Stuff\xspace} 
\newcommand{\ADE}{ADE-20K\xspace} 
\newcommand{\CITY}{Cityscapes\xspace} 
\newcommand{\IOUMF}{mIoU$_\textrm{MF}$}

\title{Unlocking Pre-trained Image Backbones for Semantic Image Synthesis}

\crefname{section}{Section}{Secs.}
\Crefname{section}{Section}{Sections}
\crefname{table}{Table}{Tabs.}
\Crefname{table}{Table}{Tables}
\crefname{figure}{Figure}{Figs.}
\Crefname{figure}{Figure}{Figures}
\crefname{appendix}{Appendix}{App.}
\Crefname{appendix}{Appendix}{App.}

\author[1,2]{Tariq Berrada}
\author[1]{Jakob Verbeek}
\author[1]{Camille Couprie}
\author[2]{Karteek Alahari}

\affiliation[1]{FAIR, Meta}
\affiliation[2]{Univ.\ Grenoble Alpes, Inria, CNRS, Grenoble INP, LJK}

\abstract{

Semantic image synthesis, i.e., generating images  from  user-provided semantic label maps,  is an important conditional image generation task as it allows to control both the content as well as the spatial layout of generated images.
Although  diffusion models have pushed the state of the art in generative image modeling, the iterative nature of their inference process makes them computationally demanding.
Other approaches such as GANs are more efficient as they only need a single feed-forward pass for generation, but 
the image quality tends to suffer on large and diverse datasets.
In this work, we propose a new class of GAN discriminators for semantic image synthesis that generates highly realistic images by exploiting feature backbone networks pre-trained for tasks such as image classification. 
We also introduce a new generator architecture with better context modeling and using cross-attention to inject noise into latent variables, leading to more diverse generated images.
Our model, which we dub \ours, achieves state-of-the-art results in terms of image quality and consistency with the input label maps on \ADE, \COCO, and \CITY,  surpassing recent diffusion models while requiring two orders of magnitude less compute for inference.
}

\date{\today}
\correspondence{Tariq Berrada (\email{tariqberrada@meta.com}) and Jakob Verbeek  (\email{jjverbeek@meta.com})}

\begin{document}

\maketitle

\begin{figure*}
\begin{center}
    \includegraphics[width=\linewidth]{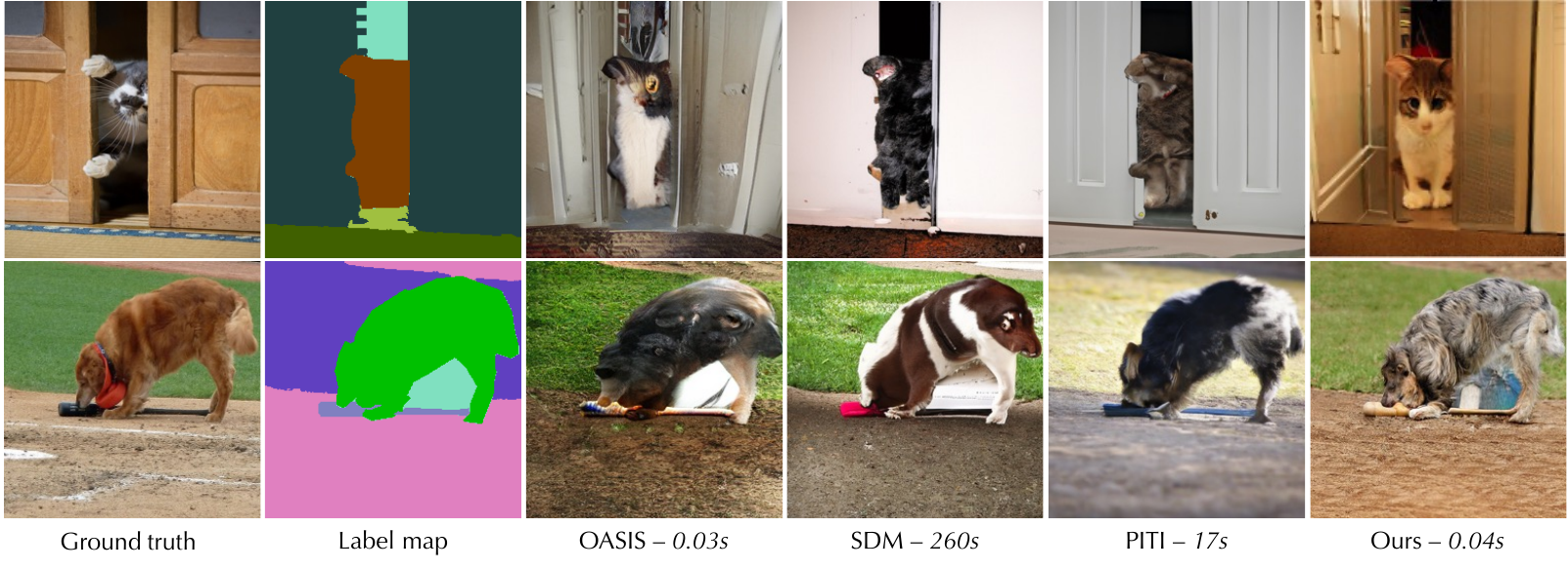}
    \caption{Images generated with  models trained on \COCO, comparing our approach to   state-of-the-art methods OASIS, SDM, and PITI, along with inference times  to generate a single image.
    Our approach combines high-quality samples with low-latency sampling.
    }
    \label{fig:teaser}
\end{center}
\end{figure*}

\section{Introduction}
\label{sec:intro}




Conditional image synthesis aims to generate images based on 
information such as text, categories, sketches, label maps, \etc.
While text-based generation has seen impressive advances in recent years with diffusion models~\citep{ldm, glide}, it lacks precise control over the location and the boundaries of objects, which are important properties for creative content generation tasks like photo editing, inpainting, and for data augmentation in discriminative learning~\citep{zhao2023xpaste, akrout2023diffusionbased, azizi2023synthetic, hemmat2023feedbackguided}.
Consequently, in this work we focus on semantic image synthesis~\citep{pix2pix2017,wang2018pix2pixHD, park2019SPADE, oasis, sdm, piti}, where the goal is to produce an image, given a segmentation map, with every pixel assigned to a category, as input. 
Because of the one-to-many nature of the mapping, prior works have tackled this problem under a conditional GAN~\citep{goodfellow2014generative} framework by exploring different conditioning mechanisms in GANs to do stochastic generations that correspond to the input label map~\citep{pix2pix2017,wang2018pix2pixHD,park2019SPADE}.
Others developed conditional discriminator models, which avoid image-to-image reconstruction losses that compromise diversity in generated images~\citep{oasis}. 
Diffusion models~\citep{sdm,piti} have also been investigated for this problem. SDM~\citep{sdm} learns to incorporate spatially adaptive normalization layers into the diffusion process of a latent diffusion model, while PITI~\citep{piti} finetunes pre-trained text-to-image diffusion models such as \citep{glide}.
In comparison to GANs, diffusion models often result in improved image quality, but suffer from lower consistency with the input segmentation maps, and are slower during inference due to the iterative denoising process~\citep{careil23cvpr}.

To improve the image quality and consistency of GAN-based approaches, we explore the use of pre-trained image backbones in discriminators for semantic image synthesis. 
Although leveraging pre-trained image models is common in  many other  vision tasks, such as classification, segmentation, or detection, and more recently for class-conditional GANs~\citep{projected_gans}, 
to our knowledge this has not been explored for semantic image synthesis. 
To this end, we develop a UNet-like encoder-decoder architecture where the encoder is a fixed pre-trained image backbone, which leverages the multi-scale feature representations embedded therein, and the decoder is a convolutional residual network.
We also propose a novel generator architecture, building on the dual-pyramid modulation approach~\citep{dp_gan} with an improved label map encoding through attention mechanisms for better diversity and global coherence among the images generated.
Finally, we introduce novel loss and regularization terms, including a contrastive loss and a diversity constraint, which improve generation quality and diversity. 

We validate our contributions with experiments on the \ADE, \COCO, and \CITY datasets. 
Our model, termed {\it \ours} for ``Discriminator Pretraining for Semantic IMage Synthesis'', achieves state-of-the-art performance in terms of image quality (measured by FID) and consistency with the input segmentation masks (measured by mIoU) across all three datasets.
Our results not only surpass recent diffusion models on both metrics, but also come with two orders of magnitude faster inference.

\noindent
In summary, 
 our main contributions are the following:
\begin{itemize}
    \item We develop an encoder-decoder discriminator that leverages feature representations from pre-trained  networks.
    \item We propose a generator architecture using attention mechanisms for noise injection and context modeling.
    \item We introduce additional loss and regularization terms that improve sample quality and diversity.
    \item We outperform state-of-the-art  GAN  and  diffusion-based methods in image quality, input consistency, and speed.
\end{itemize}

\section{Related work}
\label{sec:related}

\mypar{Generative image modeling}
Several frameworks have been explored in deep generative modeling, including GANs~\citep{goodfellow2014generative,stylegan3,  pix2pix2017, park2019SPADE, oasis, kang2023gigagan, poe_gan}, VAEs~\citep{kingma14iclr, vahdat20arxiv, vq-vae2}, flow-based models~\citep{density-nvp, kingma2018glow, fan2022styleflow} and diffusion-based models~\citep{dhariwal21nips, ho20nips, ldm, sdm, piti}.
GANs consist of generator and discriminator networks which partake in a mini-max game that  results in the generator learning to model the target data distribution. 
GANs realized a leap in sample  quality, due to the mode-seeking rather than mode-covering nature of their objective function~\citep{nowozin16nips,lucas19nips}.
More recently, breakthrough results in image quality have been obtained using  text-conditioned diffusion  models trained on large-scale text-image datasets~\citep{gafni22arxiv,ramesh22dalle2,ldm,saharia22nips,glide}. 
The relatively low sampling speed of diffusion models has triggered research on scaling GANs to training on large-scale datasets 
to achieve competitive image quality while being orders of magnitude faster to sample~\citep{kang2023gigagan}. 

\begin{figure*}
    \centering
    \includegraphics[width=\linewidth]{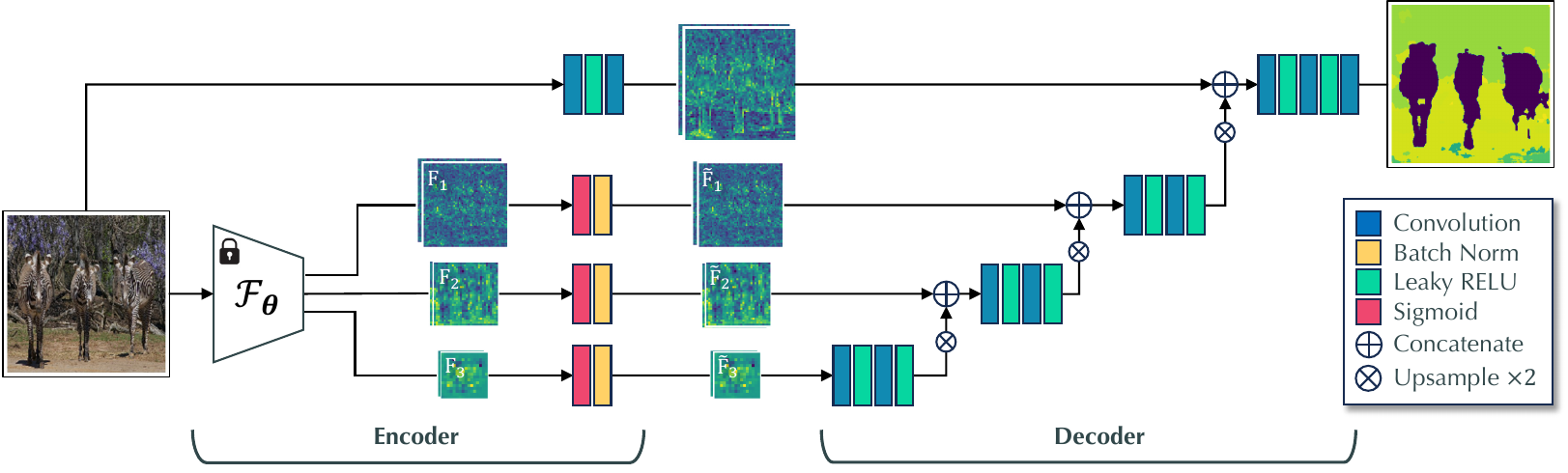}
    \caption{{\bf Architecture of our discriminator model.} The encoder consists of a pre-trained  feature backbone $\mathcal{F}_\theta$ (left), residual blocks at  full image resolution (top), and trained feature decoder that aggregates the multi-scale features from the frozen  backbone (right).
    \vspace{-4mm}
    }
    \label{fig:discriminator}
\end{figure*}

\mypar{Semantic image synthesis} 
Early approaches for semantic image synthesis leveraged cycle-consistency between generated images and  conditioning masks~\citep{pix2pix2017, wang2018pix2pixHD} and spatially adaptive normalization  (SPADE) layers~\citep{park2019SPADE}. 
These approaches combined adversarial losses with image-to-image feature-space reconstruction losses to enforce image quality as well as consistency with the input mask~\citep{zhang18cvpr}. 
OASIS~\citep{oasis} uses a UNet discriminator model which labels pixels in real and generated images with semantic classes and an additional  ``fake'' class, which overcomes the need for feature-space losses that inherently limit sample diversity, while also improving consistency with the input segmentation maps.
Further improvements have been made by adopting losses to learn image details at varying scales~\citep{dp_gan}, or through
multi-modal approaches such as PoE-GAN~\citep{poe_gan}, which leverage data from different modalities like text, sketches and segmentations. 

Several works have explored diffusion models for semantic image synthesis. 
SPADE layers  were incorporated in the denoising network of a diffusion model in SDM~\citep{sdm} to align the generated images with semantic input maps.
PITI~\citep{piti} replaced the text-encoder of pre-trained text-to-image diffusion models, with a label map encoder, and fine-tuned the resulting model. 
In our work, rather than relying on generative pre-training as in PITI,  we leverage discriminative pre-training.
Another line of work considers generating images from segmentation maps with free-text annotations~\citep{couairon23iccv,avrahami23cvpr}.
These diffusion approaches, however, suffer from poor consistency with the input label maps while also being much more computationally demanding during inference than their GAN-based counterparts.

\mypar{Pre-trained backbones in GANs}
Pre-trained feature representations have been explored in various ways in GAN training.
When the model is conditioned on detailed inputs, such as sketches or segmentation maps,  pre-trained backbones are used to define a reconstruction loss between the generated and training images~\citep{zhang18cvpr}.
Another line of work leverages these backbones as fixed encoders in adversarial discriminators~\citep{projected_gans, enhancing}.
Naively using a pre-trained encoder with a fixed decoder yields suboptimal results, thus the Projected GANs model~\citep{projected_gans} uses a feature conditioning strategy based on random projections to make the adversarial game more balanced. 
While this approach is successful with some backbones, the method worked best with small pre-trained models such as EfficientNets~\citep{efficientnets}, while larger models resulted in lower performance. 
A related line of work~\citep{ensembling_models} uses an ensemble of multiple pre-trained backbones to obtain a set of discriminators from which a subset is selected at every step for computing the most informative gradients.
This produced impressive results but has the following significant overheads which make it inefficient: (i) all the discriminators and their associated optimizers are stored in memory, (ii) there is a pre-inference step to quantify the suitability of each discriminator for any given batch, and (iii) the main discriminator is trained from scratch. 
Our work is closely related to Projected GANs, but to our knowledge the first one to leverage pre-trained discriminative feature networks for semantic image synthesis.

\mypar{Attention in GANs} While most of the popular GAN frameworks such as the StyleGAN family relied exclusively on convolutions~\citep{stylegan, Karras2019stylegan2, stylegan3}, some other works explored the usage of attention in GANs to introduce a non-local parametrization that operates beyond the receptive field of the convolutions in the form of self attention~\citep{biggan, vitgan, zhang2019selfattention, ganformer, kang2023gigagan}, as well as cross-attention to incroporate information from different modalities (text-to-image).
To the best of our knowledge, this is the first work to explore the usage of cross-attention for semantic image synthesis.
\section{Method}
\label{sec:method}

Semantic image synthesis aims to produce realistic RGB images ${\bf g}\in\mathbb R^{W \times H\times 3}$ that are diverse and  consistent  with an input label map ${\bf t}\in\mathbb R^{W \times H \times C}$, where $C$ is the number of semantic classes and $W\times H$ is the spatial resolution. 
A one-to-many mapping is ensured  by conditioning on a random noise vector ${\bf z} \sim \mathcal{N}({\bf 0, I})$ of dimension $d_z$.


In this section, we present our GAN-based approach, starting with our method to leverage pre-trained feature backbones in the discriminator (\cref{sec:backbones}). We then describe our noise injection and label map modulation mechanisms for the generator  (\cref{sec:generator}), and detail the losses we use to train our models (\cref{sec:training}).


\subsection{Pre-trained discriminator backbones}
\label{sec:backbones}
We use a UNet architecture~\citep{ronneberger15miccai} for the discriminator, similar to OASIS~\citep{oasis}. 
The discriminator is trained to classify pixels in real training images with the corresponding ground-truth labels, and the pixels in generated images as ``fake''. 
Note that the discriminator in~\citep{oasis} is trained from scratch and does not benefit from any pre-training.

Let $\mathcal{F}_\theta$ be a pre-trained feature backbone with parameters $\theta$. 
We use this backbone, frozen, as part of the ``encoder'' in the UNet discriminator. 
Let ${\bf F}_l \in\mathbb{R}^{C_l \times W_l \times H_l}$ denote the  features extracted by the backbone at levels $l=1,\dots,L$, which generally have different spatial resolutions $W_l\times H_l$ and number of channels $C_l$.
These features are then processed by the UNet ``decoder'', which is used to predict per-pixel labels spanning the semantic categories present in the input label map, as well as the ``fake'' label. 
Additionally, to exploit high-frequency details in the image, we add a fully trainable path at the full image resolution  with two relatively shallow residual blocks.
The full discriminator architecture is illustrated in \cref{fig:discriminator}. 

\mypar{Feature conditioning}
An important problem with using pre-trained backbones is feature conditioning. Typical backbones are ill-conditioned, meaning that some features are much more prominent than others. This makes it difficult to fully exploit the learned feature representation of the backbone as strong features overwhelm the discriminator's decoder and result in exploring only certain regions in the feature representation of the encoder.
Previously, \citep{projected_gans} tried to alleviate this problem by applying cross-channel mixing (CCM) and cross-scale mixing (CSM) to the features, while~\citep{ensembling_models} average the signals from multiple discriminators to obtain a more diluted signal. 
Empirically, the first approach underperforms in many of our experiments, as the strong features still tend to mask out their weaker, yet potentially relevant, counterparts. On the other hand, the second introduces a large overhead from the multiple models being incorporated in training.
In our work, we develop a method that better exploits the feature representation from the encoder. We achieve this by aiming to make all features have a comparable contribution to the downstream task.

Consider a feature map ${\bf F}_l \in\mathbb{R}^{C_l \times W_l \times H_l}$  at scale $l$  from the pre-trained backbone.
First, we apply a contractive non-linearity (CNL) such as sigmoid to obtain ${\bf F}'_l = \sigma ( {\bf F}_l )$.
Next, we normalize the features to ensure they have a similar contribution in the following layers. 
We choose batch normalization, yielding $\tilde{\bf F}_l = ({\bf F}'_l - \mu_l) / \sigma_l$, where $\mu_l$ and $\sigma_l$ are the batch statistics.
In this manner, all features are in a similar range and therefore the decoder does not prioritize features with a high variance or  amplitude. 

\subsection{Generator architecture}
\label{sec:generator}

\begin{figure}
    \centering
    \includegraphics[width=\linewidth]{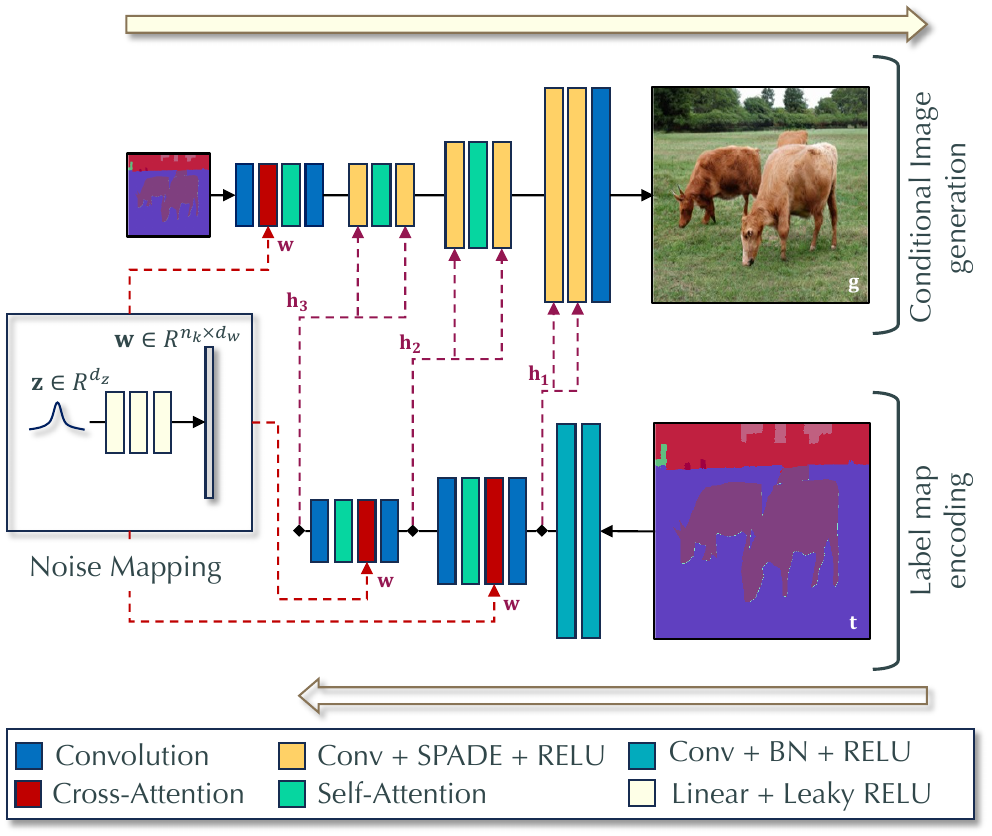}
    \caption{{\bf Our generator architecture consist of two components.}
    (i) A conditional image generation network (top) that takes a low-resolution  label  map as input and produces the full-resolution output image. 
    (ii) A semantic map encoding network (bottom) that takes the full resolution label map as input and produces  multi-scale features that are used to modulate the intermediate features of the image generation network.
    }
    \label{fig:generator}
    \vspace{-3mm}
\end{figure}
Our generator architecture is based on  DP-GAN~\citep{dp_gan}, but offers two main novelties: a revisited noise injection mechanism and improved modeling of long-range dependencies through  self-attention.
Following DP-GAN, we use a mask encoding network to condition the SPADE blocks, rather than conditioning the SPADE blocks on the label maps via a single convolution layer, which cannot take into account longer-range dependencies encoded in the label map. 
Each block of the label map encoding pyramid is made of a single convolution layer with downsampling followed by batch norm, GELU activation~\citep{GELU}, attention modules, and a pointwise convolution layer. For every scale, we obtain a modulation map ${\bf h}_i, i \in \{  1,\dots, L\}$ which, concatenated with a resized version of the ultimate map ${\bf h}_L$, will serve as conditioning for the SPADE block at the same resolution.

While \citep{oasis} argued that concatenating a spatial noise map to the label map was enough to induce variety in the generated images, since the noise is present in all SPADE blocks, and therefore hard to ignore, the same cannot be said for the architecture of DP-GAN~\citep{dp_gan}. 
The noise is injected only at the first layer of the label map encoding network, hence it is much easier to ignore.
Consequently, we propose a different mechanism for noise injection, making use of cross-attention between the learned representations at the different scales and the mapping noise obtained by feeding ${\bf z}$ to a three-layer MLP, ${\bf w} = \text{MLP}({\bf z}) \in \mathbb{R}^{n_k\times d_w}$. 
Let ${\bf h}_i \in \mathbb{R}^{C_i \times H_i \times W_i}$ be the downsampled feature representation from the previous scale, ${\bf h_i}$ first goes through a convolution to provide an embedding of the label map, then the spatial dimensions are flattened and projected via a linear layer to obtain the queries $Q \in \mathbb{R}^{H_iW_i \times d_q}$. 
The transformed  noise vector ${\bf w}$ is projected via two linear layers to obtain the keys and the values $K, V \in \mathbb{R}^{n_k \times d_q}$, then the cross-attention is computed as:
\begin{equation}
    {\bf A} = \text{SoftMax}\left( \frac{Q K^\top}{ \sqrt{d_q}} \right) V.
\end{equation}
Each noise injection block at lower spatial resolutions (\mbox{$64\times 64$} and lower) uses a noise injection module made of a residual cross-attention block
\begin{equation}
    a({\bf h}_i, {\bf w}) = {\bf h_i} + \eta_i \cdot {\bf A}({\bf h_i}, {\bf w}),
\end{equation}
where $\eta_i \in \mathbb{R}$ is a trainable gating parameter initialized at 0.
The noise injection is followed by a residual self-attention block, before having a convolution output the conditioning at scale $i$. For higher resolutions where attention modules are too costly, we use convolutional blocks only.
A diagram of the generator architecture is provided in \cref{fig:generator}. 


\subsection{Training}
\label{sec:training}





We train our models by minimizing a weighted average of three  loss functions which we detail below.

\mypar{Pixel-wise focal loss} 
Our main loss is based on a pixel-wise GAN loss~\citep{oasis}, where the discriminator aims to assign pixels in real images to the corresponding class in the conditioning label map, and those in generated images to an additional ``fake'' class. 
To improve the performance on rare classes, we replace the weighted cross-entropy of~\citep{oasis} with a weighted focal loss~\citep{lin2017focal}, while keeping the same weighting scheme as in \citep{oasis}.
Let $p({\bf x}) \in [0,1]^{H\times W\times (C+1)}$ denote the output class probability map of the discriminator for a real RGB image $\bf x$, and  $p({\bf g}) \in [0,1]^{ H\times W\times (C+1)}$ be the probability map for a generated image ${\bf g}=G({\bf z, t})$, where the label index $C+1$ is used for the ``fake'' class.
Then, the discriminator loss is:
\begin{eqnarray}
\mathcal{L}_\textrm{D}= -\mathbb{E}_{({\bf x,t})}  \sum_{c=1}^{C}\alpha_c\sum_{i=1}^{H \times W} {\bf t}_{i,c} \left( 1 - p({\bf x})_{i,c} \right)^\gamma \log p({\bf x})_{i,c}  \nonumber \\
    - \mathbb{E}_{({\bf g, t})}  \sum_{i=1}^{H \times W} \left( 1 - p({\bf g})_{i,C+1} \right)^\gamma \log p({\bf g})_{i,C+1},
\end{eqnarray}
where the  $\alpha_c$ are the class weighting terms and $\gamma$ is a hyper-parameter of the focal loss. 
The standard cross-entropy is recovered for $\gamma=0$, and for $\gamma>0$ the loss puts more weight on poorly predicted labels.

\noindent
The pixel-wise loss for the generator is then takes the form:
\begin{equation}
    \mathcal{L}_\textrm{G} = - \mathbb{E}_{(\bf g, t)}  \sum_{c=1}^C \alpha_c \sum_{i=1}^{H \times W} {\bf t}_{i,c} \left( 1 - p({\bf g})_{i,c} \right)^\gamma \log p({\bf g})_{i,c}.  
\end{equation}
Using the focal loss, both the generator and discriminator put more emphasis on pixels that are incorrectly classified, these  often belong to rare classes which helps to improve performance for these under-represented classes.
\noindent
To prevent the discriminator output probabilities from saturating and thus leading to vanishing gradients, we apply one-sided label smoothing~\citep{improved_techniques} by setting  the cross-entropy targets to $1-\epsilon$ for the discriminator loss, where $\epsilon$ is a hyper-parameter. 

\mypar{Contrastive loss} 
We define a  patch-wise contrastive loss that encourages the generated images to be globally coherent.
Our contrastive framework is based on InfoNCE~\citep{info-nce} which aims to bring matching patch features closer together, and push them further from non-matching features. 
Given a pair $({\bf x,t})$ of image and  label map, we generate a corresponding image ${\bf g}= G({\bf z}, {\bf t})$,  and use $\bf H_x$ and $\bf H_g$ the corresponding  multi-scale features obtained from a pre-trained VGG network~\citep{vgg}.
For every scale, we sample matching features $\bf z, z^+$  from the same spatial coordinates  in $\bf H_g$ and $\bf H_x$ respectively. 
Additionally, we sample $N$ non-matching features ${\bf z}^-_n$ at randomly selected coordinates from  $\bf H_x$.

The features are then projected to an embedding space using a convolution followed by a two-layer MLP to obtain ${\bf v, v^+}, {\bf v}^-_n \in \mathbb{R}^{d_v}$ before computing the InfoNCE loss as
\begin{equation}
    \mathcal{L}_\textrm{NCE}({\bf v, v^+\ccc{,} v^-}) = - \log \left( 
    \frac{e^{ \bf v^\top v^+ / \tau}}  
    {e^{ \bf v^\top v^+ / \tau} + \sum_{n=1}^N e^{{\bf v^\top} {\bf v}^-_n / \tau}} 
    \right),
\end{equation}
where $\tau$ is a temperature parameter controlling the sharpness in the response of the loss.
We apply the loss at feature scales $1/4, 1/8, 1/16$, and take their sum.
This is similar to the contrastive losses used for image-to-image translation~\citep{park2020contrastive}, and the main difference comes from the feature representation from which the loss is calculated. While other methods reuse the encoder features from their translation network, we obtain the feature pyramid from a VGG network and process these features by a simple module made of a convolution block followed by a projection MLP.



\mypar{Diversity loss}
To promote diversity among the generated images we introduce a loss that encourages two images generated with the same mask, but different latents $\bf z$, to be sufficiently distinct from each other.
In particular, we define
\begin{equation}
    \mathcal{L}_\textrm{Div}=\left[  \tau_\text{div}- \frac{ \left\lVert \sigma \circ G^{f}( {\bf z}_1, {\bf t}) -  \sigma \circ G^{f}({\bf z}_2, {\bf t}) \right\rVert_1}{\lVert {\bf z}_1 - {\bf z}_2 \rVert_1} \right]^+,
\end{equation}
where  $[\cdot]^+=\max(0,\cdot)$ retains the non-negative part of its argument, $G^{f}$ is the output of the generator before the final convolution and $\sigma$ the sigmoid function.
We adopt a cutoff threshold $\tau_\text{div}$ on the loss in order to not overly constrain the generator, and apply this loss only for similar samples given the same label map.

\begin{table*}
    \centering
    \rowcolors{4}{white}{gray!15}
\ifcvpr
\else
\resizebox{\linewidth}{!}{
\scriptsize
\begin{tabular}{lrccrccccc}
         \toprule
          & \multicolumn{3}{c}{COCO} & \multicolumn{3}{c}{ADE20k} & \multicolumn{3}{c}{Cityscapes} \\
         \cmidrule(lr){2-4}\cmidrule(lr){5-7}\cmidrule(lr){8-10}
          & FID ($\downarrow$) & mIoU$_\textrm{MF}$ ($\uparrow$) & mIoU ($\uparrow$) & FID ($\downarrow$) & mIoU$_\textrm{MF}$ ($\uparrow$) & mIoU ($\uparrow$) & FID ($\downarrow$) & mIoU$_\textrm{MF}$ ($\uparrow$) & mIoU ($\uparrow$)\\
         \midrule
         Pix2pixHD~\citep{wang2018pix2pixHD} & 111.5 & --- & 14.6 & 73.3 & --- & 22.4 & 104.7 & --- & 52.4 \\
         SPADE~\citep{park2019SPADE} & 22.6 & --- & 37.4 & 33.9 & --- & 38.5 & 71.8 & --- & 62.3  \\
         OASIS~\citep{oasis} & 17.0 & 52.1 & 44.1  & 28.3 & 53.5 & 48.8 & 47.7 & 72.0 & 69.3\\
         DP-GAN~\citep{dp_gan} & --- & --- & --- & 26.1 & --- & 52.7 & 44.1 & --- & 73.6 \\
         PoE-GAN~\citep{poe_gan} & 15.8 & --- & --- & --- & --- & --- & --- & --- & ---\\
         \midrule
         SDM~\citep{sdm} & 15.9 & 40.3 & 36.8  & 27.5 & 51.9 & 44.0  & 42.1 & 72.8 & 69.1\\
         {PITI}~\citep{piti} & 15.5 & 31.2 & 29.5 & --- & --- & --- & --- & --- & ---\\
         \midrule
         \ours (ours) &  \bf 13.6 & \bf 65.2 & \bf 57.7 & \bf 22.7 & \bf 67.8 & \bf 54.3 & \bf 38.2 & \bf 78.5 & \bf 76.3 \\
         \bottomrule
    \end{tabular} }
\fi
    \caption{Comparison of our results to the state-of-the-art GAN-based (first block) and diffusion-based (second block) methods. 
    Scores for these methods are taken from their corresponding papers.
    We also report mIOU scores with Mask2Former (mIoU$_\textrm{MF}$) for methods where pre-trained checkpoints or generated images are available.
    }
    \vspace{-3mm}
    \label{tab:eval_main}
\end{table*}

\section{Experiments}
\label{sec:exp}

We present our experimental setup in \cref{sec:setup}, followed by our main results in \cref{sec:results}, and ablations in \cref{sec:ablations}.

\begin{table}
    \centering
    \rowcolors{2}{white}{gray!15}
\ifcvpr
\else
\resizebox{\columnwidth}{!}{%
\scriptsize
\setlength{\tabcolsep}{4pt}
\begin{tabular}{lrrrrr}
        \toprule
        Backbone & Prms.\  & FLOPS & Acc@1\ & FID ($\downarrow$) & mIoU$_\textrm{MF}$ ($\uparrow$)\\
        \midrule
        Swin-B & 107 & 15.4G & \bf 86.4 & 29.5 & 55.4\\
        ResNet-50 & 44 & 4.1G & 76.2 & 24.6 & 60.5\\
        EfficientNet-Lite1 & 3 & 631M & 83.4 & 24.5 & 63.1\\
        ConvNeXt-B & 89 & 15.4G & 85.1 & 23.5 & 63.5\\
        ConvNeXt-L & 198 & 34.4G & 85.5 & \bf 22.7 & \bf 67.8\\
        \bottomrule
    \end{tabular}
}
\fi
    \caption{Comparison of different backbone architectures on \ADE (ordered by FID). 
   The number of parameters (Prms.) is reported in millions.
   For each backbone, we also report the ImageNet-1k top-1 accuracy (Acc@1) for reference.
    }
    \label{tab:backbone_archi}
    \vspace{-2mm}
\end{table}

\subsection{Experimental setup}
\label{sec:setup}

\mypar{Datasets}
We consider three popular datasets to benchmark semantic image synthesis: \COCO, \CITY, \ADE. 
\COCO  provides 118k training images and 5k  validation images, labeled with 183 classes. 
\CITY  contains 2,975 training images along with a validation set of 500 images, and uses 35 labels.
 \ADE holds 26k images with object segmentations across 151 classes. 
 Similar to \citep{park2019SPADE,sdm,wang2018pix2pixHD}, we use instance-level annotations when available. 
For \COCO and \CITY, we use instance segmentations as in~\citep{cheng2020panoptic}, by creating vertical and horizontal offset maps of every foreground pixel \wrt it's object center of mass, and concatenate these to the semantic label maps as input for the model. 
For \ADE there are no instance segmentations available.
We generate images at a resolution of $256 \times 256$ for \ADE and \COCO, and $256\times 512$ for \CITY.
In order to avoid training models on identifiable personal data, we use  anonymized versions of the datasets; see the supplementary material for more details. 

\mypar{Metrics}
We compute FID~\citep{heusel17nips} to assess image quality. 
We also report the mean intersection-over-union score (mIoU) to measure the consistency of the generated image with the input segmentation maps.
For a fair comparison with previous work~\citep{oasis,dp_gan,park2019SPADE}, we used the segmentation models from these works for inferring label maps of the generated images: UperNet101~\citep{xiao18eccv} for \ADE, multi-scale DRN-D-105~\citep{yu17cvpr} for \CITY, and DeepLabV2~\citep{chen18pami} for \COCO. We refer to the scores obtained with these models as mIoU.
In addition, we infer label masks using Mask2Former~\citep{cheng2021mask2former}, which is more accurate than other segmentation models, thus yielding a more meaningful comparison to the ground-truth masks. We denote the resulting scores as mIoU$_\textrm{MF}$. 
Additional details are provided in the supplementary material.

\mypar{Implementation details} 
We counter the strong class imbalance in the datasets used in our experiments  with a  sampling scheme favoring rare classes. 
Let $f_c$ be the fraction of training images where class $c$ appears, then each image is sampled with a probability proportional to $f_k^{-1/2}$ with $k$ the sparsest class present in the image. 

Each of our models is trained on one or two machines with eight V100 GPUs. 
We set the total batch size at $64$ and use ADAM optimizer in all our experiments with a learning rate of $10^{-3}$ and momentums $\beta_1\!=\!0, \beta_2\!=\!0.99$. 
For pre-trained Swin backbones, we found it necessary to use gradient clipping to stabilize training. 
Following prior work, we track an exponential moving average of the generator weight and set the decay rate to $\alpha=0.9999$. 
For the contrastive loss, we set the weighting factor $\lambda_{C}=100$, the temperature $\tau=0.3$ and select $N=128$ negative samples.
We set $\lambda_\text{GAN}=1$ for the GAN loss and $\lambda_D=10$ for the diversity loss. 
For the focal loss, we set the focusing parameter $\gamma=2$. 

\begin{figure*}
    \centering
    \includegraphics[width=\linewidth]{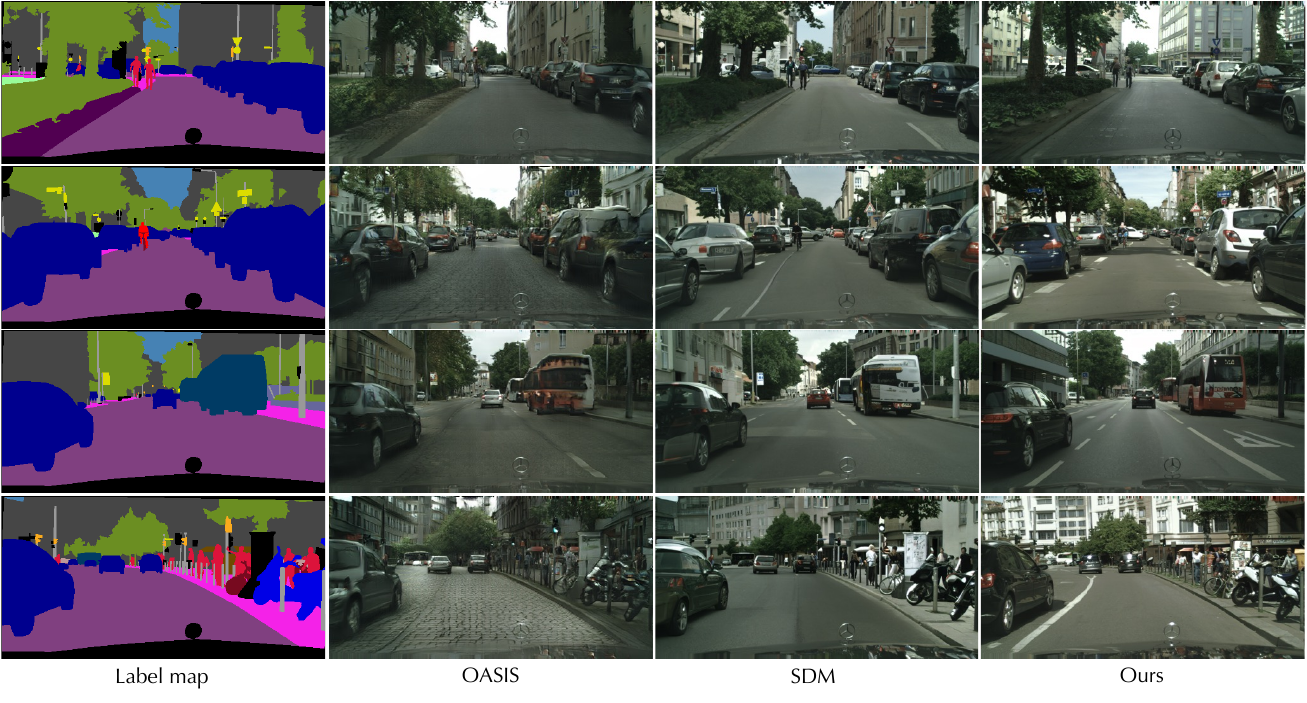}
    \vspace{-7mm}
    \caption{Qualitative comparison with prior work on the \CITY dataset. We show the results of OASIS~\citep{oasis}, SDM~\citep{sdm}, and our approach along with the corresponding label map used for generating each image. Note that our method generates more coherent objects with  realistic textures in comparison.
    \vspace{-3mm}
    }
    \label{fig:cityscapes_comparison}
\end{figure*}

\subsection{Main results}
\label{sec:results}

\mypar{Comparison to the state of the art}
In \cref{tab:eval_main}, we report the results obtained with our model in comparison to the state of the art. 
Our \ours method (with ConvNext-L backbone) achieves the best performance across metrics and datasets.
On \COCO, we improve the FID  of 15.5 from PITI to 13.6, while improving the \IOUMF~ of 52.1 from OASIS to 65.2.
For \ADE, we observe a similar trend with an improvement of 3.4 FID points \wrt DP-GAN, and an improvement of 14.5 points in \IOUMF ~\wrt OASIS.
For \CITY we obtain smaller improvements of 3.9 FID points \wrt Semantic Diffusion and 5.5 points in  \IOUMF.
See \cref{fig:teaser} and \cref{fig:cityscapes_comparison} for qualitative comparisons of model trained on  \COCO and \CITY. 
Please refer to the supplementary material for additional examples, including ones for \ADE.



\begin{table}
        \centering
        \rowcolors{2}{white}{gray!15}
    {\scriptsize
        \begin{tabular}{lccc}
            \toprule
            Pre-training & Acc@1 &  FID ($\downarrow$) & mIoU$_\textrm{MF}$ ($\uparrow$)\\
            \midrule
            IN-1k@224 & 84.3 & \bf 22.7 & 62.8\\
            IN-21k@224  & 86.6 & 23.6 & 64.1\\
            IN-21k@384 & 87.5 & \bf 22.7 & \bf 67.8\\
            \bottomrule
        \end{tabular}}
        \caption{Influence of discriminator pre-training on the overall performance for \ADE using a ConvNext-L  backbone.
        }
        \vspace{-3mm}
        \label{tab:pre-training}
\end{table}

\mypar{Encoder architecture} 
We experiment with different pre-trained backbone architectures for the discriminator in \cref{tab:backbone_archi}. All the encoders were trained for ImageNet-1k classification.
We find that the attention-based Swin architecture~\citep{liu2021Swin} has the best ImageNet accuracy, but compared to convolutional models performs worse as a discriminator backbone for semantic image synthesis, and tends to be more unstable, often requiring gradient clipping to converge. 
For the convolutional models, better classification accuracy translates to better FID and \IOUMF.

\mypar{Pre-training dataset}
In \cref{tab:pre-training}, we analyze the impact of pre-training the  ConvNext-L architecture in  different ways and training our models on top of these, with everything else being equal. 
We consider pre-training on ImageNet-1k (IN-1k@224) and ImageNet-21k (IN-21k@224) at $224\times 224$ resolution, and also on ImageNet-21k at $384\times 384$ resolution (IN-21k@384).
In terms of \IOUMF, the results are in line with those observed for different architectures:  discriminators trained with larger datasets (IN-21k) and on higher resolutions perform the best.
On the other hand, we find that for FID, using the standard ImageNet (IN-1k@224) results in better performance than its bigger IN-21k@224 counterpart, and performs as well as IN-21k@384 pre-training.
This is likely due to the use of the same dataset in the Inception model~\citep{szegedy16cvpr}, which is the base for calculating FID, introducing a bias in the metric.


\begin{table}
    \centering
    \rowcolors{2}{white}{gray!15}
    \ifcvpr
\else
\resizebox{\columnwidth}{!}{%
\scriptsize
\begin{tabular}{lrrrrr}
        \toprule
        Model & Gen.\ steps & Ups.\ steps & $\Delta t_\text{gen} $ & $\Delta t_\text{ups} $ & $\Delta t_\text{tot} $\\
        \midrule
        PITI & 250 & 27 & 14.3 & 3.1 & 17.4~~ \\
        PITI & 27 & 27 & 1.5 & 3.1& 4.6~~ \\
        SDM & 1000 & --- & 260.0 & --- & 260.0~~ \\
        \midrule
        \ours & --- & --- & --- & --- & 0.04\\
        \bottomrule
    \end{tabular}
}
\fi
    \caption{Comparison of inference time (in seconds) of PITI, SDM and our GAN-based model.
    We show the time taken by the generative ($\Delta t_\text{gen}$) and the upsampling ($\Delta t_\text{ups}$) models in addition to the total time ($\Delta t_\text{tot}$) for these steps.
    }
    \vspace{-4mm}
    \label{tab:inf_speed}
\end{table}

\mypar{Inference speed} 
An important advantage of GAN models over their diffusion counterparts is their fast inference. 
While a GAN only needs one forward pass to generate an image, a diffusion model requires several iterative denoising steps, resulting in slower inference, which can hamper the practical usability of the model.
In \cref{tab:inf_speed}  we report the inference speed for generating a single $256 \times 256$ image, averaged over $50$ different runs. 
PITI uses 250 denoising steps for the generative model at $64\times64$ resolution and 27 steps for the upsampling model by default, while SDM uses 1000 steps at full resolution.
We also benchmark using 27 steps for the PITI generative model.
Our generator is two to three orders of magnitude faster than its diffusion counterparts.

\ifcvpr
\else 
\begin{table}
{
\scriptsize
\begin{center}
     \begin{tabular}{lcc}
        \toprule
         & FID ($\downarrow$) & mIoU$_\textrm{MF}$ ($\uparrow$) \\
        \midrule
            \rowcolor{gray!15}
            Baseline - no normalization & 27.8 & 58.6\\
            Projected GAN & 28.9 & 59.1\\
                    \rowcolor{gray!15}
            \ours w/o sigmoid & 24.9 & 62.7\\ 
             \ours w/o BatchNorm & 26.0 & 61.6\\
            \rowcolor{gray!15}\ours & \bf 24.5 & \bf 63.1\\
        \bottomrule
    \end{tabular}
\end{center}
}
 \caption{Ablation on feature conditioning shown on \ADE with EfficientNet-Lite1 backbone.
    }
    \label{tab:cond_abl}
\end{table}
\fi

\subsection{Ablations}
\label{sec:ablations}

\mypar{Feature Conditioning}
We perform an ablation to validate our  feature conditioning mechanisms on \ADE in \cref{tab:cond_abl}.
We compare the normalization in \ours to: (i) a baseline approach where the backbone features are unchanged before entering the discriminator network (``Baseline - no normalization''), (ii) the Projected GAN~\citep{projected_gans} approach with cross-channel and scale mixing, and using the normalization layer without (iii) the contractive non-linearity (``/ours w/o sigmoid'') or (iv) normalisation (``\ours w/o BatchNorm''). 
For a fair comparison with \citep{projected_gans}, these experiments are conducted on their best reported backbone, EfficientNet-Lite1~\citep{efficientnets}.
Compared to the baseline method, the Projected GAN approach improves \IOUMF~by 0.5 points, but degrades FID by 1.1 points.  
Using our feature conditioning based on BatchNorm and sigmoid activation, we improve the \IOUMF~ by 4.5 points and FID by 3.3 points \wrt the baseline.
When we leave out the sigmoid activation both \IOUMF~ and FID drop by 0.4 points. We observe a similar behavior without normalization: FID and \IOUMF~ both decrease by 1.5 points.
These results validate the fittingness of our feature conditioning approach.

\mypar{Architectural modifications}
In \cref{tab:gen_abl}, we perform an ablation on our proposed architectural modifications. 
Swapping out our generator or discriminator with the ones from  OASIS, suggests that most of the gains are due to our discriminator design. Using the OASIS discriminator instead of ours deteriorates \IOUMF by 18.8 points and FID by 6.6 points.
We also experiment with removing the cross-attention noise injection mechanism and replacing it with the usual concatenation instead, as well as leaving out the self-attention layers.
Both of these contribute to the final performance in a notable manner.
Finally, we include an ablation on label smoothing, which deteriorates FID by 0.3 and \IOUMF by 1.4 points when left out.

\ifcvpr
\else
\begin{table}
    \centering
    {\scriptsize
\begin{tabular}{lcc}
        \toprule
         & FID ($\downarrow$) & mIoU$_\textrm{MF}$ ($\uparrow$) \\
        \midrule
        \rowcolor{gray!15}
        \ours & \bf 22.7 & \bf 67.8\\ 
        \midrule
        \multicolumn{3}{c}{Generator architecture}\\
        \midrule
        \rowcolor{gray!15}
        OASIS disc + our gen & 29.3 & 49.0\\
        OASIS gen + our disc & 25.6 & 63.6\\
        \rowcolor{gray!15}
        Ours w/o self-attention & 23.7 & 65.4\\
        Ours w/o cross-attention & 23.6 & 64.5\\
        \midrule
        \multicolumn{3}{c}{Training}\\
        \midrule
        \rowcolor{gray!15}
        Ours w/o label smoothing & 23.0 & 66.3\\
        Ours w/o contrastive loss & 25.1 & 66.0\\
        \bottomrule
    \end{tabular}
    }
    \caption{Ablations on the architectural design and training losses, shown on \ADE with ConvNext-L backbone. 
    }
    \label{tab:gen_abl}
\end{table}
\fi

\mypar{Contrastive loss}
To assess the importance of the contrastive loss, we perform an ablation in the last row of \cref{tab:gen_abl} where we remove it during training. 
This substantially impacts the results: worsening FID by 2.4 and mIoU$_\textrm{MF}$ by 1.8 points.
In \cref{tab:contr_abl} we evaluate different values for the temperature parameter $\tau$. 
We find an optimal temperature parameter $\tau_C = 0.3$, using $\lambda_c=100$.

\ifcvpr
\else
\begin{table}
    \centering
    {\scriptsize
        \rowcolors{2}{white}{gray!15}
\begin{tabular}{lcccc}
        \toprule
         $\tau$ & 0.07 & 0.3 &0.7 & 2.0\\
         \midrule
         FID & 25.7 & {\bf 22.7} & 24.1 & 26.4\\
         \IOUMF & 62.6 & {\bf 67.8} & 66.3 & 61.4\\
         \bottomrule
    \end{tabular}
    }
    \caption{Influence of the contrastive loss evaluated on \ADE. 
    }
    \label{tab:contr_abl}
\end{table}
\fi

\mypar{Focal loss}
In \cref{tab:pixelwise_loss}, we consider the impact of the focal loss by comparing it to the weighted cross-entropy loss, as used in OASIS, and the effect of class weighting in the focal loss. 
This comparison is conducted on \CITY because it is more imbalanced than \ADE.
We find that switching from weighted cross-entropy to the focal loss improves FID by 0.5 points but worsens \IOUMF~by 0.9 points. When comparing weighted focal loss to weighted CE, we observe that FID improves by 1.6 points, and \IOUMF~by 2.6 points.

\ifcvpr
\else
\begin{table}
    \centering
    {\scriptsize
    \rowcolors{2}{white}{gray!15}
    \begin{tabular}{lcc}
        \toprule
        Loss & FID ($\downarrow$) & \IOUMF ($\uparrow$)\\
        \midrule
        Weighted CE & 39.8 & 75.9\\
        Focal & 39.3 & 75.0\\
        Weighted Focal  & {\bf 38.2} & {\bf 78.5}\\
        \bottomrule
    \end{tabular}
    }
    \caption{Comparison of pixel-wise losses on \CITY with ConvNext-L backbone.}
    \label{tab:pixelwise_loss}
\end{table}
\fi

\begin{figure}
    \centering
    \includegraphics[width=\columnwidth]{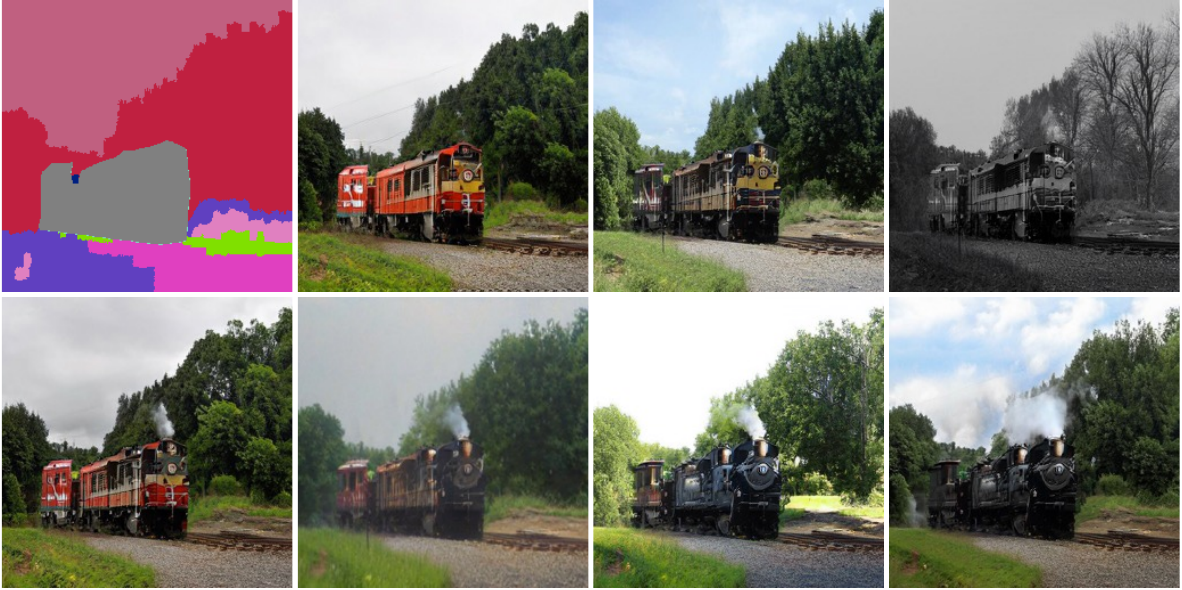}
    \caption{Images generated  by varying the noise vector with \ours trained on \COCO and   using a ConvNext-L backbone.}
    \label{fig:variablity}
    \vspace{-3mm}
\end{figure}

\mypar{Diversity}
We investigate the effect of the diversity loss on the variability of  generated images. 
Following~\citep{oasis}, we report the mean LPIPS distance across 20 synthetic images from the same label map,  averaged across the validation set, in \cref{fig:diveristy_abl}.  
A qualitative example is provided in \cref{fig:variablity} showing a clear variety in the images generated.
In comparison with OASIS, we generate more diverse images, with an LPIPS score similar to that of SPADE, but with a much higher quality, as reported in \cref{tab:eval_main}, in terms of FID and \IOUMF.

\ifcvpr
\else
\begin{table}
    \centering
    {\scriptsize
    \rowcolors{2}{white}{gray!15}
    \begin{tabular}{lcc}
        \toprule
        Model & 3D noise & LPIPS ($\uparrow$) \\
        \midrule
        SPADE+& \cmark & 0.16\\
        SPADE+ & \xmark & 0.50\\
        OASIS & \cmark & 0.35\\
        \midrule
        \ours & \xmark & 0.47\\
        \bottomrule
    \end{tabular}
    }
    \caption{Evaluation of the diversity of images generated.
    Results for SPADE+ and OASIS are taken from~\citep{oasis}.
    }
    \label{fig:diveristy_abl}
\end{table}
\fi

\section{Conclusion} \label{conclusion}
We introduced \ours that harnesses pre-trained   backbones in GAN-based semantic image synthesis models. 
We achieve this by using them as an encoder in UNet-type discriminators, and introduce a feature conditioning approach to maximize the effectiveness of the pre-trained features.
Moreover, we propose a novel generator architecture which uses cross-attention to inject noise in the image generation process, and introduce new loss terms to boost sample diversity and input consistency.
We experimentally validate our approach and compare it to state-of-the-art prior work based on GANs as well as diffusion models on three standard benchmark datasets. Compared to these, we find improved performance in terms of image quality, sample diversity, and consistency with the input segmentation maps. 
Importantly, with our approach inference is two orders of magnitude faster than diffusion-based approaches.


In our experiments we found that transformer-based models, such as Swin, can lead to instability  when used as discriminator backbones.
Given their strong performance for dense prediction tasks, it would be worthwhile to further study  and mitigate this issue in future work, hopefully bringing additional improvements.


{
    \small
    \bibliographystyle{ieeenat_fullname}
    \bibliography{jjv,main}
}

\clearpage
\appendix
\clearpage

\setcounter{figure}{0} 
\renewcommand\thefigure{S\arabic{figure}} 
\setcounter{table}{0}
\renewcommand{\thetable}{S\arabic{table}}

\section{More details on the experimental setup}
\label{sec:more_setup}

\subsection{Architecture details}
\label{sec:arch_details}

\begin{figure*}
    \centering
    \includegraphics[width=.8\linewidth]{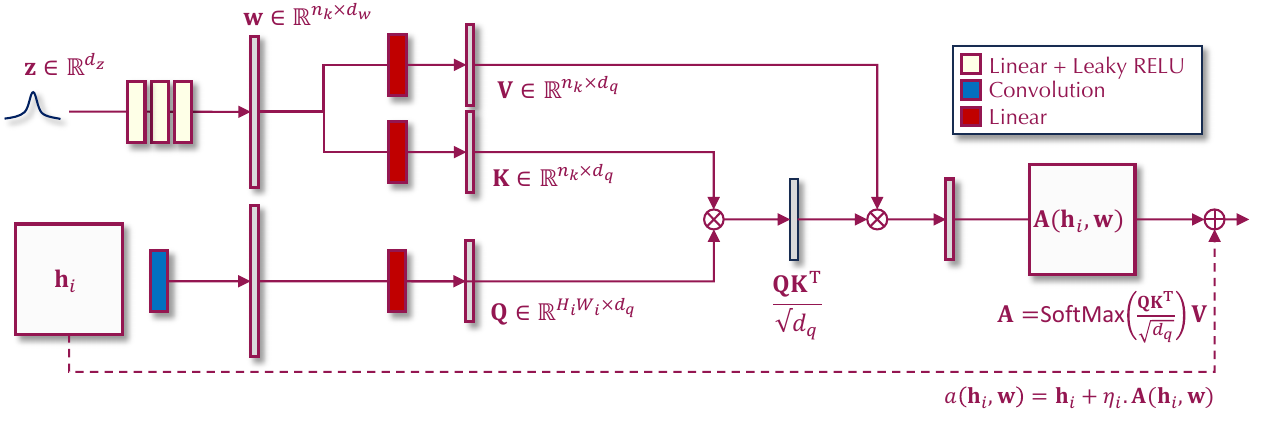}
    \caption{Schematic overview of our noise injection mechanism using cross-attention.}
    \label{fig:cross_att}
\end{figure*}

\mypar{Generator}
As can be seen in Figure~3 
of the main paper, our generator consists of a UNet-like architecture with two pyramidal paths.
The {\it label map encoding} takes the input segmentation map, and progressively downsamples it to produce label-conditioned multi-scale features.
These features are then used in the {\it image generation path}, which progresively upsamples the signal to eventually produce an RGB image.
The stochasticity of the images generated is based on conditioning on the noise vector ${\bf z}$. We provide a schematic overview of the noise injection operation in \Cref{fig:cross_att}.
In \Cref{tab:gen_details}, we provide additional information on the label map encoding and the image generation paths.

In the label map encoding branch, each block is made of the elements listed in \Cref{tab:gen_details}. Cross-attention and self-attention are only applied at lower resolutions ($64 \times 64$, and lower) where we use an embedding dimension that is half of the original feature dimension. 
We downscale the feature maps by using convolution layers with a stride of 2.

In the image synthesis branch, we follow the same architecture as OASIS~\cite{oasis} with the only difference being the SPADE conditioning maps which are given by the label map encoding path instead of a resized version of the label maps.
We also remove the hyperbolic tangent at the output of the network as we found it leads to a more stable generator.

For the contrastive learning branch, features obtained from VGG19 go through three convolutional blocks and two linear layers for projection.
We sample $128$ different patches to obtain negative samples from the image.






\begin{table*}
    \centering
    \scriptsize
    \rowcolors{0}{white}{gray!15}
    \begin{tabular}{lc}
        \toprule
         Parameter & Description\\
         \midrule
         \rowcolor{white}
         \multicolumn{2}{c}{{\bf Hyperparameters}}\\  
         \midrule
         ${\bf z}$ dimension & 64\\
         ${\bf w}$ dimension & 256\\
         Batch size & 64\\
         Learning rate & $10^{-3}$\\
         $\beta_1$ for Adam & 0\\
         $\beta_2$ for Adam & 0.99\\
         EMA beta & 0.9999\\
         \midrule
         \rowcolor{white}
         \multicolumn{2}{c}{{\bf Label map encoding}}\\
         \midrule
         \rowcolor{gray!15}
         Pyramid block & Conv2d(kernel\_size=3), BN, GELU, CrossAttention, BN, \\
         \rowcolor{gray!15} &          GELU, SelfAttention, GELU, BN, Conv2d(kernel\_size=1)\\
         Self Attention channel divider & 2\\
         Cross Attention channel divider & 2\\
         Conv block & Conv2d(kernel\_size=3), BN, GELU, Conv2d(kernel\_size=1) \\
         Block type & [Conv, Conv, Conv, Linear, Linear]\\
         \midrule
         \rowcolor{white}
         \multicolumn{2}{c}{{\bf Image synthesis branch}}\\
         \midrule
         Channel base & 64\\
         Number of residual blocks & 6\\
         Channel depths & [1024, 1024, 1024, 512, 256, 128, 64]\\
         Residual block & SPADE, Leaky RELU, Conv2d(3)\\
         Pyramid dimensionality & 64\\
         Hyperbolic tangent on output & No\\
         \midrule
         \rowcolor{white}
         \multicolumn{2}{c}{{\bf Contrastive learning branch}}\\
         \midrule
         Perceptual network & VGG19\\
         Contrastive encoding channels & [64, 128, 256, 512, 512]\\
         Contrastive embedding dimension & 256\\
         Number of patches & 128\\
         \bottomrule
    \end{tabular}
    \caption{Architecture details of the generator.
    }
    \label{tab:gen_details}
\end{table*}

\mypar{Discriminator}
We provide additional details of our discriminator architecture in \Cref{tab:disc_details}.
The residual blocks are made of one convolution with kernel size 3 followed by leaky ReLU, then a pointwise convolution with leaky ReLU.
For the full resolution channel, we set the dimensionality to 128.
For the lower resolution channels, we stick to the same dimensionality as the corresponding encoder feature.
The dimensionality of the final convolution before predicting the segmentations is set to 256.


We use spectral norm on all convolutional and linear layers in both the generator and the discriminator.

\begin{table*}
    \centering
    {\scriptsize
    \rowcolors{0}{white}{gray!15}
    \begin{tabular}{lc}
        \toprule
        Parameter & Description\\
         \midrule
         \rowcolor{white}
         \multicolumn{2}{c}{{\bf Hyperparameters}}\\
         \midrule
        Number of multiscale backbone features & 4\\
        Full resolution embedding dimension & 128\\
        Number of residual blocks & 5\\
        \midrule
        \rowcolor{white}
        \multicolumn{2}{c}{{\bf Decoder}}\\
        \midrule
        Residual block & Conv2d(kernel\_size=3), Leaky RELU, Conv2d(kernel\_size=1), Leaky RELU\\
        Leaky RELU slope & 0.2\\
        Penultimate channel dimension & 256\\
        \midrule
        \rowcolor{white}
        \multicolumn{2}{c}{{\bf Feature conditioning}}\\
        \midrule
        Conditioning normalization & Batch Norm w.o learned affine\\
        Conditioning non-linearity & Hyperbolic tangent\\
        
         \bottomrule
    \end{tabular}}
    \caption{Architecture details of the discriminator. 
    }
    \label{tab:disc_details}
\end{table*}

\mypar{Feature conditioning}
In ProjectedGAN's paper~\cite{projected_gans}, they observe that when using a fixed feature encoder in the GAN discriminator, only a subset of features is covered by the projector. They therefore propose propose to dilute prominent features, encouraging the discriminator to utilize all available information equally across the different scales.
However, they do not delve deeper into the practical reasons behind this.
Namely, feature encoders trained for a discriminative task will have different structures to those trained on generative tasks. For the former, models tend to capture a subset of key features while disregarding other less relevant features.
On the latter however, the model needs an extensive representation of the different objects it should generate.
In practice, this translates to feature encoders having bad conditioning. 
The range of activations differs greatly from one feature to the other, which leads to bias towards a minority features that have a high amplitude of activations.
A simple way to resolve this issue is by applying normalization these features to have a distribution with zero mean and a unit standard deviation across the batch.

In some situations, linear scaling of the features might not be enough to have proper conditioning of the features.
Accordingly, we reduce the dynamic range of the feature maps before the normalization by using a sigmoid activation at the feature outputs of the pretrained encoder.

\mypar{Fixed vs finetuned backbone}
We experimented with finetuning the whole discriminator, encoder included. In all of our attempts, such models diverged without producing meaningful results.
We therefore chose to keep the encoder fixed for this work.

\subsection{Computation of the mIoU evaluation metrics}
To compute the mIoU metric, we infer segmentation maps for  generated images.
As noted in the main paper, we infer segmentation maps for the generated images using the same networks as in OASIS~\cite{oasis}, that is:  UperNet101~\cite{xiao18eccv} for \ADE, multi-scale DRN-D-105~\cite{yu17cvpr} for \CITY, and DeepLabV2~\cite{chen18pami} for \COCO.
We also measure mIoU using Mask2Former~\cite{cheng2021mask2former} with Swin-L backbone~\cite{liu2021Swin} (\IOUMF), which infers more accurate segmentation mask, thus producing a more accurate comparison to the ground-truth masks.

In \Cref{tab:miou} we compare the segmentation accuracy on the three datasets we used in our experiments. 
The results confirm that Mask2Former is more accurate for all three datasets, in particular on \COCO, where it boosts mIoU by more than 19 points \wrt DeepLab-v2.

\begin{table}
\centering
\rowcolors{0}{white}{gray!15}
{\scriptsize
\begin{tabular}{lccc}
\toprule
& \ADE  & \CITY & \COCO \\
\midrule
UperNet101 & 42.7 & --- & --- \\
MS DRN-D-105 & --- & 61.3 & --- \\ 
DeepLab-v2  & --- & --- & 35.3 \\
\midrule
Mask2Former &  45.3  & 69.9 & 54.5\\ 
\bottomrule
\end{tabular}
}
\caption{Segmentation performance in terms of mIoU on  real images using different segmentation models.
To match the setting used in our semantic image synthesis experiments, evaluation images are downsampled to $256\times256$ for \ADE and COCO, and to $256\times512$ for \CITY.
}
\label{tab:miou}
\end{table}

\begin{table}
    \centering
    \resizebox{\linewidth}{!}{
    \scriptsize
    \begin{tabular}{llccc}
        \toprule
        Dataset & Model & Blurring & FID ($\downarrow$) & \IOUMF ($\uparrow$)\\
        \midrule
        \rowcolor{gray!15}
         & OASIS & \xmark & 17.0 & 52.1\\
        \rowcolor{gray!15}
         & OASIS & \cmark & 18.8 & 47.1\\
        \rowcolor{gray!15}
        \multirow{-3}{*}{ \COCO} & \ours (ours) &  \cmark & \bf 13.6 & \bf 65.2\\ 
         & OASIS & \xmark & 28.3 & 53.5\\
         & OASIS & \cmark & 29.1 & 49.6\\
        \multirow{-3}{*}{\ADE} & \ours (ours) & \cmark & \bf 22.7 & \bf 67.8\\
         \rowcolor{gray!15}
         & OASIS  & \xmark & 47.7 & 72.0\\
         \rowcolor{gray!15}
         & OASIS & \cmark & 48.0 & 71.6\\
         \rowcolor{gray!15}
         \multirow{-2}{*}{\CITY} & \ours (ours)& \cmark & \bf 38.2  & \bf 78.5\\
        \bottomrule
    \end{tabular}}
    \caption{Influence of face blurring on the performance of OASIS.
    }
    \label{tab:blurring}
\end{table}

\subsection{Influence of face blurring}
\label{sec:faceblur}
To avoid training models on identifiable personal data, we did not include such information in the training dataset. 
For \CITY we use the release of the dataset with blurred faces and licence plates, which is available publicly on the website listed in \Cref{tab:assets_link}.
For \ADE and \COCO we executed our pipeline to detect, segment and blur faces. 
To this end, we use Retina-Net~\cite{lin2017focal} and segment anything (SAM)~\cite{kirillov23arxiv} to detect and blur human faces.
First we detect the bounding boxes of the faces for each image in our dataset, then we obtain its corresponding segmentation through SAM.
Finally, this face region detected by SAM is blurred with a Gaussian filter. 

To assess the impact of blurring, we train OASIS on blurred images using the original source code from the authors and compare to their reported results on the non-blurred data. 
We report our results in \Cref{tab:blurring}.
Here, and elsewhere in the paper, we also use the blurred data to compute FID \wrt the generated images.
We see that blurring has a negative impact on FID, most notably for \COCO (+1.8), and to a lesser extent for \ADE (+0.8) and \CITY (+0.3). 
The \IOUMF~ scores also degrade on all the datasets when using  blurred data: on \COCO, \ADE and \CITY by 5.0, 3.9, and 0.4 points respectively.
Note that in all comparisons to the state of the art, we report metrics obtained using models trained on blurred data for our approach, and models trained on non-blurred data for other approaches. 
Therefore, the real gains of our method over OASIS (and probably other methods as well) are even larger than what is shown in our comparisons in Table 1 
in the main paper.

%


\subsection{Carbon footprint estimation} 
\label{sec:footprint}
On \COCO, it takes approximately 10 days = 240 hours to train our model using 8 GPUs. On \ADE and \CITY the training times are about 6 and 4 days respectively. 
Given a thermal design power (TDP) of the V100-32G GPU equal to 250W, a power usage effectiveness (PUE) of 1.1, a carbon intensity factor of 0.385 kg CO$_2$ per KWh, a time of 240 hours $\times$ 8 GPUs = 1920 GPU hours. The $250 \times 1.1 \times 1920$ = 528 kWh used to train the model is approximately equivalent to a CO$_2$ footprint of 528 $\times$ 0.385 = 208 kg of CO$_2$ for \COCO.
For \ADE this amounts to 124 kg of CO$_2$, and  83 kg of CO$_2$ for \CITY.

\section{Additional experimental results} 
\label{sec:add_exp}

\subsection{Quantifying the ImageNet bias}
Since our backbones are pre-trained on ImageNet which is the same dataset that Inceptionv3~\cite{szegedy16cvpr} used for the standard FID model was trained on, a natural question arises whether our results are influenced by this bias towards the ImageNet dataset.
To analyze this, we report in \cref{tab:clip_fid} a quantitative comparison following the approach outlined in \cite{ss_repr}, where we compute the Fréchet distance between two Gaussians fitted to feature representations of the SwAV Resnet50 network that was pretrained in a self-supervised manner.
Our models retain state-of-the-art performance with respect to this metric on all the three datasets studied.
Additionally, we further experiment with the influence of the backbone pre-training in \Cref{tab:swav_datasets}.
Differently from the main paper where FID is studied, we find than the IN-21k checkpoint brings about better performance than its IN-1k counterpart.
While the fine-tuning at high resolution (384 vs 224) also improves swav-FID.

\begin{table}[ht]
    \centering
    \rowcolors{0}{white}{gray!15}
    {\scriptsize
    \begin{tabular}{lccccc}
        \toprule
         & OASIS & SDM & PITI & \ours\\
        \midrule
        \COCO & 3.09 & 2.68 & 2.52 & {\bf 2.14}\\
        \ADE & 4.35 & 3.85 & --- & {\bf 2.84}\\
        \CITY & 4.75 & 3.94 & --- & {\bf 3.71}\\
        \bottomrule
    \end{tabular}}
    \caption{Evaluation of SwAV Resnet50 FID on different methods. We use ConvNext-L for our method.}
    \label{tab:clip_fid}
\end{table}

\begin{table}[ht]
    \centering
    \rowcolors{0}{white}{gray!15}
    {\scriptsize
    \begin{tabular}{lcc}
        \toprule
        Pre-training & Acc@1 & FID$_\text{SwAV}$ $(\downarrow)$\\
        \midrule
        IN-1k@224 & 84.3 & 3.03\\
        IN-21k@224 & 86.6 & 2.97\\
        IN21k@384 & 87.5 & {\bf 2.84}\\
        \bottomrule
    \end{tabular}}
    \caption{Evaluation of SwAV Resnet50 FID with different pre-trainings.}
    \label{tab:swav_datasets}
\end{table}

\begin{figure*}[ht]
    \centering
    \includegraphics[width=\linewidth]{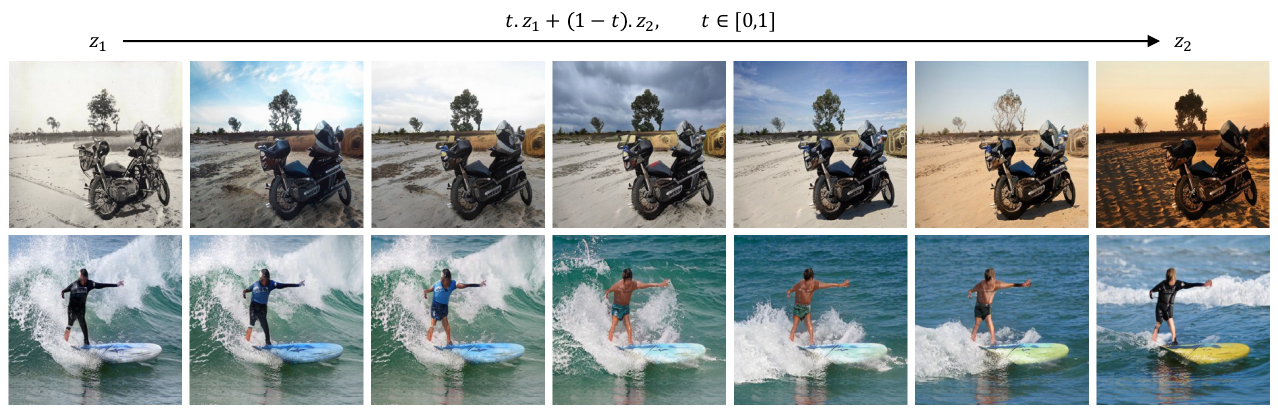}
    \caption{Noise vector interpolation. By interpolating the noise vector between two different values, we can identify the factors of variation in the image which correspond to differences in colors, textures and also object structures.}
    \label{fig:interpolation_coco}
\end{figure*}

\subsection{Influence of diversity loss}

We obtain the diversity cutoff threshold $\tau_\text{div}$ by computing the mean distance between different generated images in a batch and averaging across 5k different batches in the training set:

\begin{equation}
    \tau_\text{div} = \sum_{i, j \in \mathcal{B}} 
    \frac{\left \lVert \sigma \circ G^f(x_i, {\bf z_i}) - \sigma \circ G^f(x_j, {\bf z_j}) \right\rVert_1}{ \lVert {\bf z}_i - {\bf z}_j \rVert_1}.
\end{equation}
The distance is computed in the feature space of the generator before the final convolution that produces the image. It is then normalized by the distance between the noise vectors.
We found it necessary to add a sigmoid activation on the features before computing the loss, similar to backbone conditioning. 
Not using this activation results in unwanted effects where a tiny patch in the image has extremely high values leading to a large distance between two otherwise similar images.

\begin{figure}
    \centering
    \includegraphics[width=.9\linewidth]{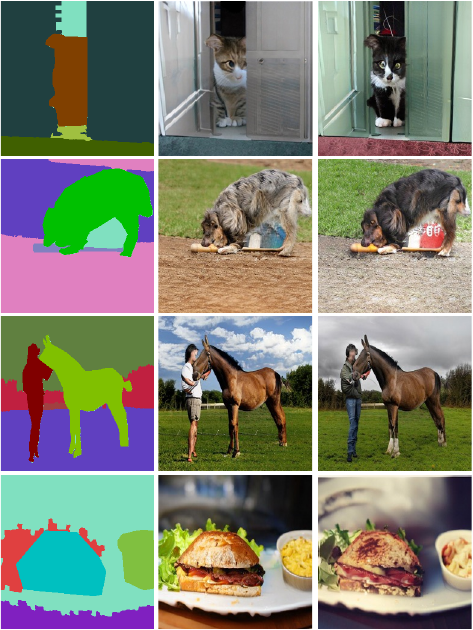}
    \caption{Additional examples of diversity in our generated images.}
    \label{fig:variability_supp}
\end{figure}

We conduct a more in-depth analysis on the impact of the diversity loss on the image quality and diversity.
We train our model with a ConvNext-L backbone  with different values for the diversity loss $\lambda_\text{div}$.
These results are reported in \Cref{tab:diversity_lambda}.
Without the diversity loss, the generator ignores the noise input, which translates to a low LPIPS score. Improving diversity with a weight of $\lambda_\text{div}=10$ results in better image quality, measured by FID as well as \IOUMF, while adding a variability factor induced from the noise input.
If the weight of the diversity loss is too high, image quality deteriorates (FID reduces by 0.6 points and \IOUMF~ diminishes by 0.1 points) without any noticeable improvements in diversity (no improvements in LPIPS by setting $\lambda_\text{div}=100$ instead of 10).
We provide more examples of diverse generations in \Cref{fig:variability_supp}, and in \Cref{fig:interpolation_coco} we perform noise vector interpolation in order to better illustrate the variation factors induced by the injected noise.

\begin{table}[ht]
    \centering
    {\scriptsize
        \rowcolors{2}{white}{gray!15}
        \begin{tabular}{lccc}
        \toprule
        $\lambda_\text{div}$ & 0 & 10 & 100\\
        \midrule
        FID $(\downarrow)$ & 22.9  & 22.7 & 23.3 \\
        \IOUMF $(\uparrow)$& 67.7  & 67.8 & 67.7 \\
        LPIPS $(\uparrow)$& 1.5e-5 & 0.47 & 0.36\\
        \bottomrule
    \end{tabular}}
    \caption{Influence of diversity loss weight on model performance. We evaluate image quality using FID and \IOUMF~ metrics while diversity is evaluated using LPIPS.}
    \label{tab:diversity_lambda}
\end{table}


\subsection{Sampling strategy}
\begin{figure*}
    \centering
   \includegraphics[width=\linewidth]{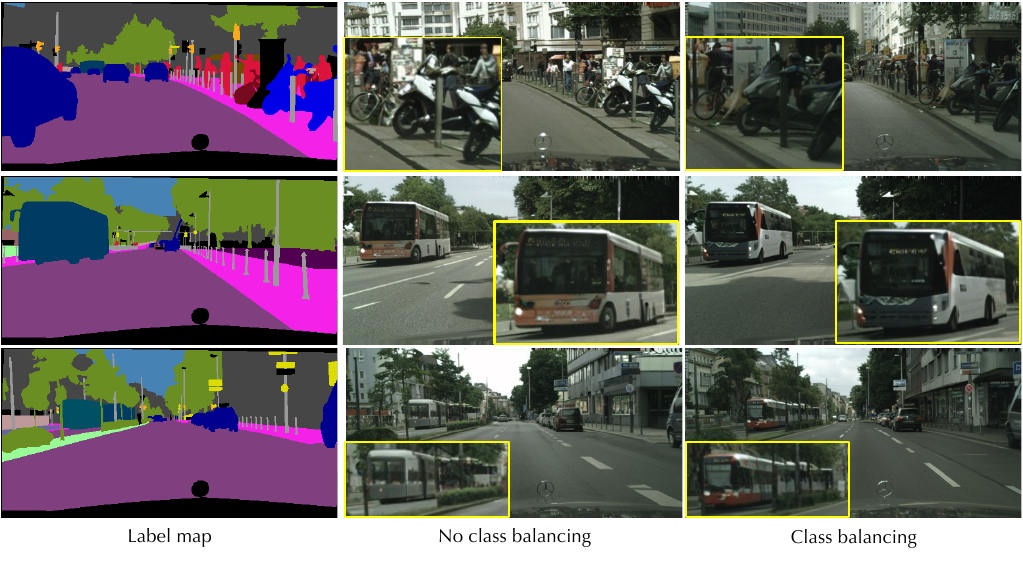}
    \caption{Qualitative examples of images generated with and without balanced sampling to train models on \CITY. 
    }
    \label{fig:sampling-fig}
\end{figure*}

We quantify the influence of the balanced sampling strategy with respect to standard uniform sampling on \COCO and \CITY datasets.
We report these results in \Cref{tab:sampling}, and find that  balanced sampling yields performance gains in both FID and mIoU for both the datasets.
In \Cref{fig:sampling-fig}, we present qualitative examples of images generated with the model trained on \CITY. 
Balanced sampling clearly leads to improvements in the visual quality of objects such as scooters, buses and trams.

\begin{table}[ht]
    \centering
    {\scriptsize
    \begin{tabular}{llcc}
        \toprule
        Dataset & Sampling strategy & FID ($\downarrow$) & \IOUMF ($\uparrow$)\\
        \midrule
        \rowcolor{gray!15}
         & Uniform  & 14.1 & 62.9\\
        \rowcolor{gray!15}
        \multirow{-2}{*}{\it \COCO} & Balanced & {\bf 13.6} & {\bf 65.2}\\
         & Uniform & 38.7 & 75.6\\
        \multirow{-2}{*}{\it \CITY} & Balanced & {\bf 38.3} & {\bf 78.3}\\
        \bottomrule
    \end{tabular}}
    \caption{Influence of sampling strategy for models trained on the \COCO and \CITY datasets.
    }
    \label{tab:sampling}
\end{table}

\begin{figure*}
    \centering
    \includegraphics[width=\linewidth]{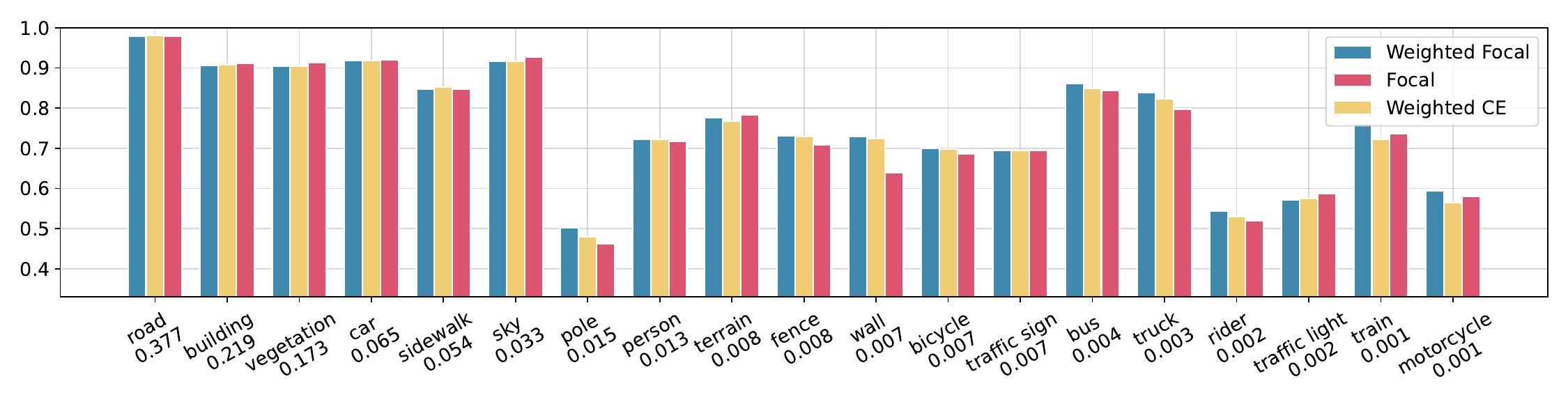}
     \includegraphics[width=\linewidth]{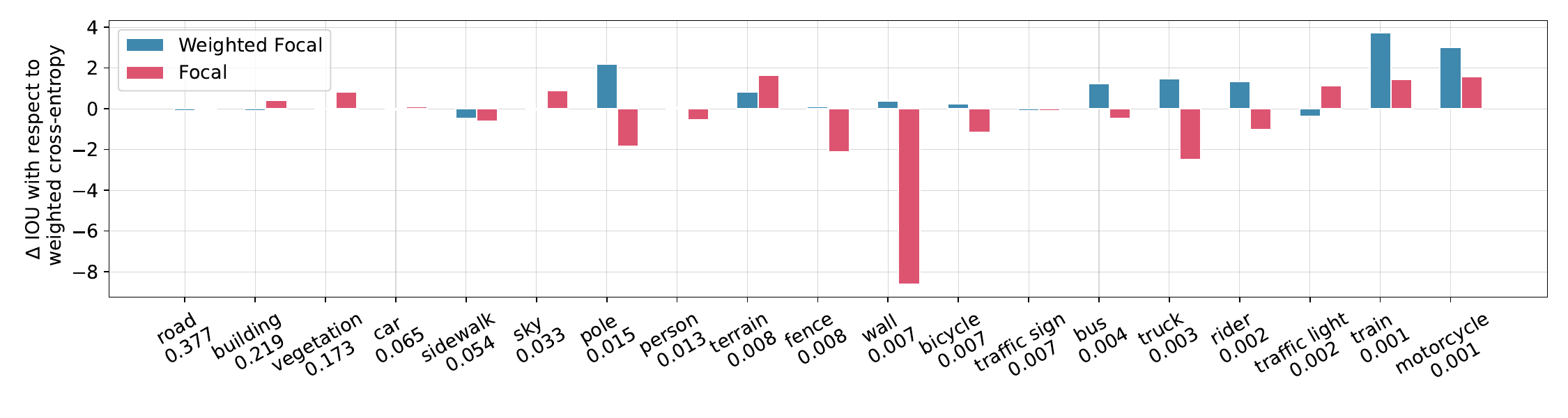}
    \caption{Top: Per-class IOU$_\textrm{MF}$ on Cityscapes with models trained with different loss functions using ConvNext-L backbone. Labels are sorted according to their frequency in the validation images, which is written below the class name.
    Bottom: Per-class difference in IOU$_\textrm{MF}$ of models trained with weighted and non-weighted focal loss \wrt the model trained with weighted cross-entropy (CE) loss. (Best viewed in color.)
    }
    \label{fig:classwise_iou}
\end{figure*}

\subsection{Influence of pixel-wise loss function}
In \Cref{fig:classwise_iou}, we compare the per-class mIoU values when training using different loss functions: weighted cross-entropy (as in OASIS), focal loss, and weighted focal loss.
This extends the class-aggregated results reported in Table 8 
in the main paper. 
These experiments were conducted on the \CITY dataset using a pre-trained ConvNext-L backbone for the discriminator.
Our use of the weighted focal loss to train the discriminator results in improved IoU for most classes. The improvements tend to be larger for rare classes. Class weighting is still important, as can be seen from the deteriorated IoU for a number of classes when using the un-weighted focal loss.

\subsection{Influence of instance-level annotations}
Since some works do not use the instance masks \cite{oasis, dp_gan, piti}, we provide an additional ablation where we train our models on \COCO and \CITY without the instance masks to isolate the gains in performance  they may bring.
For both these datasets, we observe deterioration in the model's performance when not using instance masks.
The difference is less noticeable on \COCO where the labels are already partially separated, FID only increases by 0.3 points.
On the other hand, this difference is more acute in \CITY where FID increases by 1.9 points while \IOUMF reduces by 2.2 points.
In \CITY, instances are not separated in the semantic label maps, this adds more ambiguity to the labels presented to the model which makes it more difficult to interpret them in a plausible manner.

\begin{table}[ht]
    \centering
    {\scriptsize
    \begin{tabular}{lccc}
        \toprule
        Dataset & Instance masks & FID ($\downarrow$) & \IOUMF ($\uparrow$)\\
        \midrule
        \rowcolor{gray!15}
         & \xmark &  13.9 & 65.0\\
        \rowcolor{gray!15}
        \multirow{-2}{*}{ \COCO} & \cmark & {\bf 13.6} & {\bf 65.2}\\
         & \xmark & 40.1 & 76.3\\
        \multirow{-2}{*}{\CITY} & \cmark & {\bf 38.2} & {\bf 78.5}\\
        \bottomrule
    \end{tabular}}
    \caption{Influence of instance masks on model perofrmance.}
    \label{tab:instance_mask}
\end{table}

\subsection{Larger discriminators}
For larger datasets, having huge encoder architectures could prove beneficial in capturing the complexty of the dataset.
Accordingly, we train a model on \COCO using a ConvNext-XL model.
It is approximately 1.76 times bigger than ConvNext-L we used previously, with 350M parameters.
In \Cref{tab:convnext-xl}, we report its performance as a pre-trained feature encoder in our discriminator.

\begin{table}[ht]
    \centering
    \rowcolors{0}{white}{gray!15}
    {\scriptsize
    \begin{tabular}{lcc}
         \toprule
          & FID & mIOU \\
         \midrule
         \ours (ConvNext-L) & 13.6 & 65.2\\
        \ours (ConvNext-XL) & {\bf 13.3} & {\bf 68.0}\\
        \bottomrule
    \end{tabular}}
    \caption{Performance of model trained with ConvNext-XL on \COCO.}
    \label{tab:convnext-xl}
\end{table}

\noindent
As expected, the larger ConvNext-XL encoder produces even better performance, reaching state of the art in terms of both FID and mIOU.
We observe a difference of 0.3 points in FID and 2.8 points in \IOUMF.

\subsection{Qualitative samples} \label{sec:quali}
\begin{figure*}[ht]
    \centering
    \includegraphics[width=.9\linewidth]{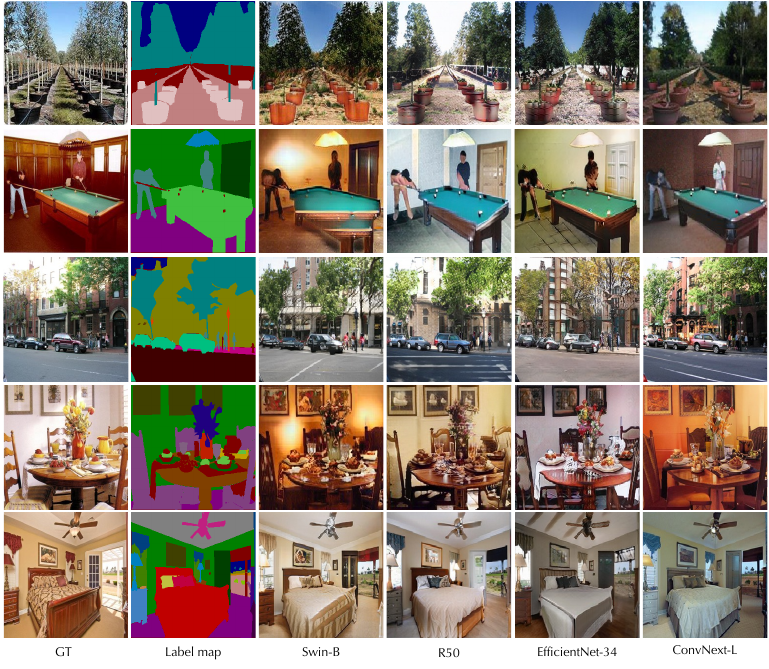}
    \caption{Qualitative comparison of \ours on \ADE using  different  backbones: Swin-B, Resnet50 (R50), EfficientNet-34,  and ConvNext-L.
    }
    \label{fig:encoders}
\end{figure*}

\begin{figure*}[ht]
    \centering
    \includegraphics[width=.9\linewidth]{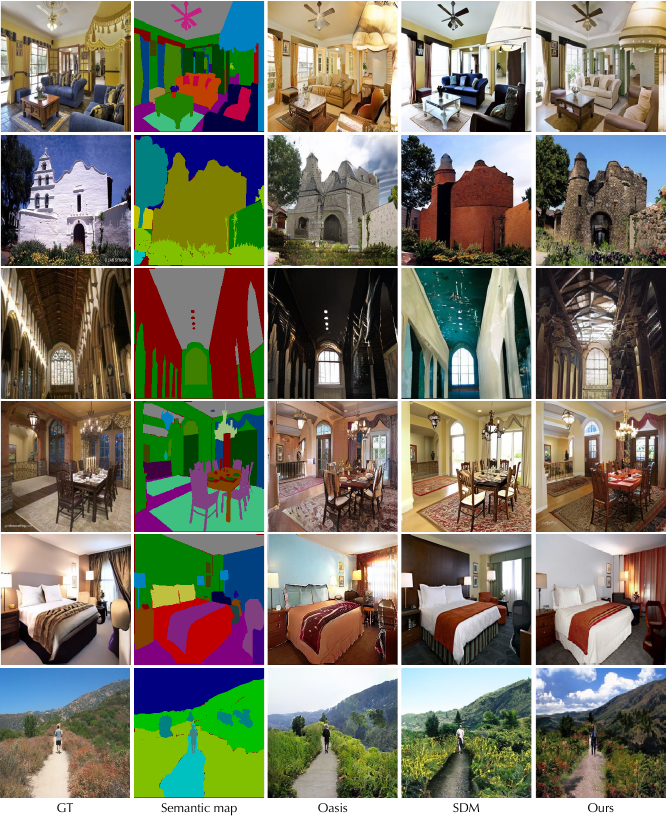}
    \caption{Qualitative comparison with prior work on \ADE, using a ConvNext-L backbone for \ours (Ours).}
    \label{fig:comparison_ade20k}
\end{figure*}

\begin{figure*}[ht]
    \centering
    \includegraphics[width=.9\linewidth]{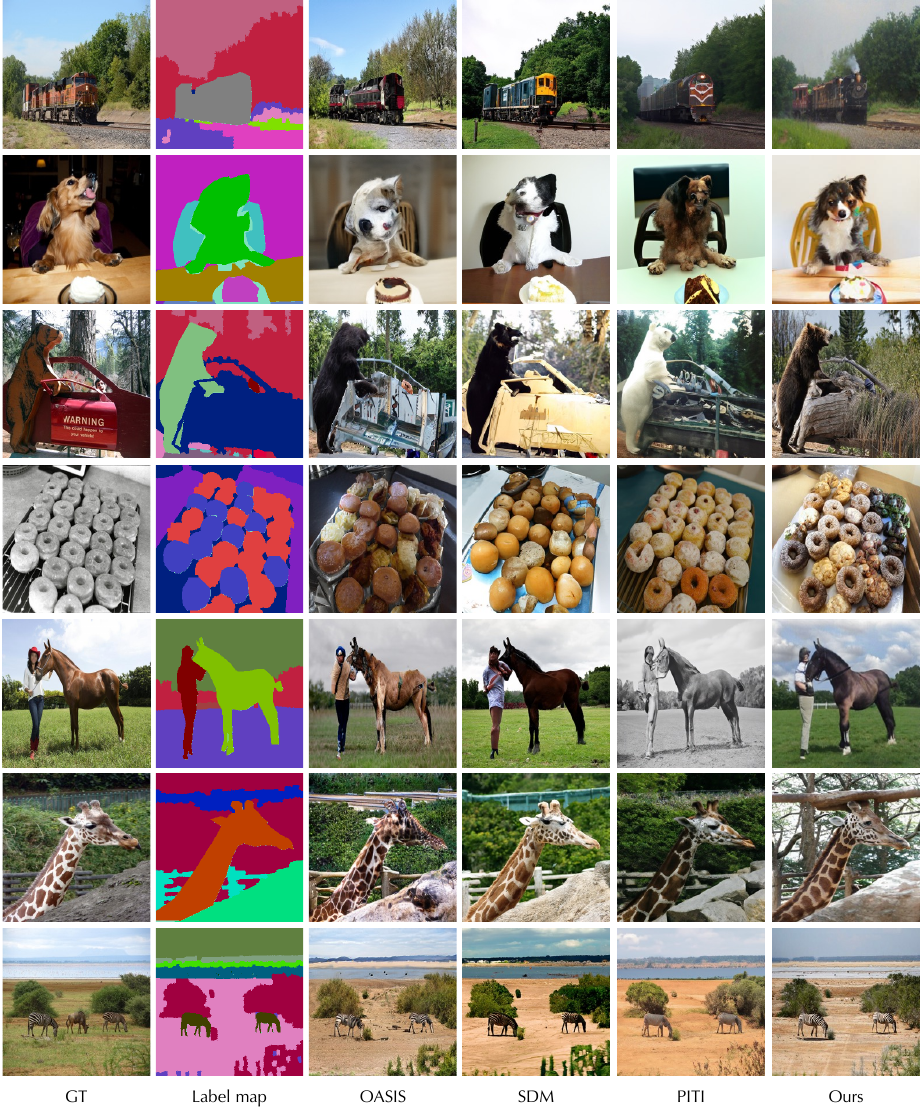}
    \caption{Qualitative comparison with prior work on \COCO, using a ConvNext-L backbone for \ours (Ours).}
    \label{fig:coco_examples}
\end{figure*}

\begin{figure*}[ht]
    \centering
    \includegraphics[width=\linewidth]{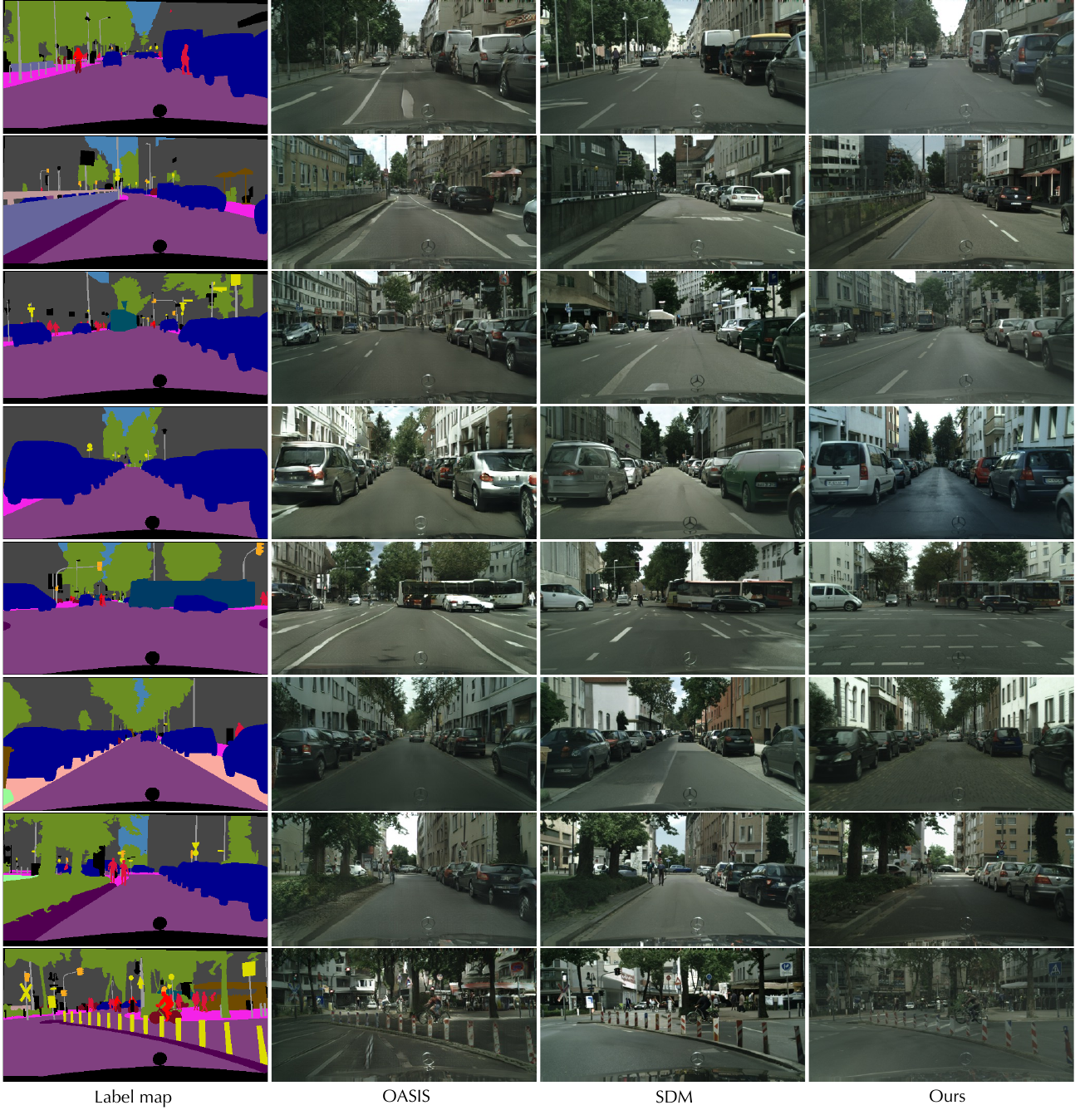}
    \caption{Qualitative comparison with prior work on \CITY, using a ConvNext-L backbone for \ours (Ours).}
    \label{fig:city_examples}
\end{figure*}

We provide qualitative samples of the images generated with our \ours model using different pre-trained backbones for the discriminator in \Cref{fig:encoders}.
In \Cref{fig:comparison_ade20k}, \Cref{fig:coco_examples}, and \Cref{fig:city_examples} we provide examples of images generated with our \ours model and compare to other state-of-the-art models on the \ADE, \COCO, and \CITY datasets, respectively.

\section{Assets and licensing information}
\label{sec:assets}
In \Cref{tab:assets_link} and \Cref{tab:assets_licensing}, we provide the links to the datasets and models used in or work  and their licensing.



\begin{table*}[t]
\small
    \begin{subtable}[b]{\linewidth}
        \centering
        {\scriptsize
        \rowcolors{0}{white}{gray!15}
        \begin{tabular}{lc}
        \toprule
            Name & Link \\
            \midrule
            ImageNet & \url{https://www.image-net.org}\\
            \COCO & \url{https://cocodataset.org}\\
            \CITY & \url{https://www.cityscapes-dataset.com}\\
            \ADE & \url{https://groups.csail.mit.edu/vision/datasets/ADE20K/}\\
            Detectron2 & \url{https://github.com/facebookresearch/detectron2}\\
            ConvNext & \url{https://github.com/facebookresearch/ConvNeXt}\\
            Swin & \url{https://github.com/microsoft/Swin-Transformer}\\
            EfficientNet & \url{https://github.com/lukemelas/EfficientNet-PyTorch}\\
            VGG19 & \url{https://github.com/pytorch/vision/blob/main/torchvision/models/vgg.py}\\
            Deeplab-v2 & \url{https://github.com/kazuto1011/deeplab-pytorch/}\\
            UperNet101 & \url{https://github.com/CSAILVision/sceneparsing}\\
            MS DRN-D-105 & \url{https://github.com/fyu/drn}\\
            Mask2Former & \url{https://github.com/facebookresearch/Mask2Former}\\
            Self-supervised FID & \url{https://github.com/stanis-morozov/self-supervised-gan-eval}\\
            \bottomrule
        \end{tabular}
        }
        \caption{ Links to the assets used in the paper.}
        \label{tab:assets_link}
    \end{subtable}
    \\
    \begin{subtable}[b]{\linewidth}
        \centering
        {\scriptsize
        \rowcolors{0}{white}{gray!15}
        \begin{tabular}{lc}
            \toprule
            Name & License \\
            \midrule
            ImageNet & Terms of access: \url{https://www.image-net.org/download.php}\\
            COCO-Stuff & \url{https://www.flickr.com/creativecommons}\\
            Cityscapes & \url{https://www.cityscapes-dataset.com/license}\\
            \ADE & \url{https://groups.csail.mit.edu/vision/datasets/ADE20K/terms/}\\
            Detectron2 &  Apache-2.0 license\\
            R50 & BSD\\
            ConvNext & MIT License\\
            Swin & MIT License\\
            EfficientNet & Apache-2.0 license\\
            VGG19 &  BSD-3-Clause license\\
            UperNet101 &BSD-3-Clause license \\
            MS DRN-D-105 &  BSD-3-Clause license\\
            Deeplab-v2 & MIT License\\
            Mask2Former & MIT License\\
            \bottomrule
        \end{tabular}
        }
        \caption{Assets licensing information.}
        \label{tab:assets_licensing}
    \end{subtable}
    
\caption{We provide information about the assets used in our work. Top : links to the assets used. Bottom : Licensing information of the relevant assets.}
\end{table*}

\end{document}